%% file: main.tex
\documentclass{article}

\usepackage{microtype}
\usepackage{graphicx}
\usepackage{subcaption}
\usepackage{booktabs}

\usepackage{hyperref}

\usepackage[preprint]{icml2027}

\input{preamble_icml}

\icmltitlerunning{Understanding and Mitigating Hallucination Escape in Tool-Using LLM Agents}

\begin{document}

\twocolumn[
  \icmltitle{Understanding and Mitigating Hallucination Escape\\
    in Tool-Using LLM Agents}

  \icmlsetsymbol{equal}{*}

  \begin{icmlauthorlist}
    \icmlauthor{Peigui Qi}{equal,ustc}
    \icmlauthor{Kunsheng Tang}{equal,ustc}
    \icmlauthor{Yide Song}{uw}
    \icmlauthor{Weiming Zhang}{ustc}
    \icmlauthor{Nenghai Yu}{ustc}
  \end{icmlauthorlist}

  \icmlaffiliation{ustc}{University of Science and Technology of China}
  \icmlaffiliation{uw}{University of Washington}

  \icmlcorrespondingnames{Weiming Zhang \textless{}zhangwm@ustc.edu.cn\textgreater{} and Nenghai Yu \textless{}ynh@ustc.edu.cn\textgreater{}}

  \icmlkeywords{Tool Learning, LLM Agents, Hallucination, Robustness, Interpretability}

  \vskip 0.3in
]

\printAffiliationsAndNotice{\icmlEqualContribution}

\begin{abstract}
\input{sections/abstract}
\end{abstract}

\input{sections/01_introduction}
\input{sections/02_related}
\input{sections/03_escape}
\input{sections/04_mechanism}
\input{sections/05_method}
\input{sections/06_discussion}
\input{sections/07_conclusion}

\section*{AI Disclosure Statement}
We used generative AI tools to assist with English-language editing, LaTeX
integration, and formatting of this manuscript. The experimental protocol also
uses GPT-5.6-Terra to construct tool configuration variants
(Appendix~\ref{app:a}). All AI-assisted revisions were reviewed and edited by the authors to preserve the intended scientific meaning and ensure consistency with the methods, experimental results, and conclusions. We take full responsibility for the final content of this work, including all text and artifacts produced with the assistance of generative AI.

\section*{Impact Statement}
This work studies and mitigates tool hallucination in LLM agents using publicly
available models and benchmarks. \method{} operates as an
inference-time intervention that does not modify model weights or introduce new
training data. None of the tool configurations involve real user data or API
calls. We do not foresee negative societal impacts specific to this work; on
the contrary, reducing tool hallucination contributes to more reliable
deployment of tool-using LLM agents. As agents increasingly act on real systems
whose tool interfaces keep evolving, \method{}'s robustness to such changes
can help prevent costly or irreversible errors in future deployments.

\bibliography{references}
\bibliographystyle{icml2027}

\clearpage
\appendix
\onecolumn
\AppendixLayout
\input{appendices/A}
\input{appendices/B}
\input{appendices/C}
\input{appendices/D}
\input{appendices/E}
\input{appendices/F}
\input{appendices/G}
\input{appendices/H}

\end{document}

%% file: preamble_icml.tex
\usepackage[T1]{fontenc}
\usepackage[utf8]{inputenc}
\usepackage{amsmath,amssymb,mathtools}
\usepackage{multirow}
\usepackage{wrapfig}
\usepackage{pifont}
\usepackage{tikz}
\usetikzlibrary{positioning, calc}
\usepackage{arydshln}
\usepackage{soul}
\usepackage{array,tabularx,longtable,adjustbox,needspace,xurl,ragged2e,makecell,etoolbox,colortbl}
\usepackage{placeins}

\definecolor{myred}{RGB}{200,60,60}
\definecolor{mygreen}{RGB}{40,150,80}

\newcommand{\method}{\textsc{EscapeGuard}}
\newcommand{\negdelta}[1]{\textcolor{myred}{$_{-#1}$}}
\newcommand{\posdelta}[1]{\textcolor{mygreen}{$_{+#1}$}}
\newcommand{\shline}{\noalign{\hrule height 1.0pt}}
\newcommand{\code}[1]{{\ttfamily\def\UrlBreaks{\do\.\do\_\do\-}\nolinkurl{#1}}}

\newcommand{\TableFont}{\footnotesize}

\newlength{\BlockSep}

\numberwithin{equation}{section}

\pretocmd{\subsection}{\Needspace{4\baselineskip}}{}{}
\pretocmd{\subsubsection}{\Needspace{3\baselineskip}}{}{}

\newcommand{\AppendixLayout}{%
  \renewcommand{\TableFont}{\small}%
  \numberwithin{table}{section}%
  \pretocmd{\section}{\FloatBarrier}{}{}%
  \raggedbottom
}

\hypersetup{pdfsubject={}}
\makeatletter
\newcommand{\icmlcorrespondingnames}[1]{\gdef\icml@corrnames{#1}}
\renewcommand{\printAffiliationsAndNotice}[1]{\global\icml@noticeprintedtrue%
  \stepcounter{@affiliationcounter}%
  {\let\thefootnote\relax\footnotetext{\hyphenpenalty=10000\exhyphenpenalty=10000\hspace*{-\footnotesep}#1%
      \forloop{@affilnum}{1}{\value{@affilnum} < \value{@affiliationcounter}}{%
        \ifnum\value{@affilnum}>1 \ \fi\textsuperscript{\arabic{@affilnum}}\csname @affilname\the@affilnum\endcsname
      }.%
      \ifdefined\icml@corrnames\ Correspondence to: \icml@corrnames.\fi

      \ \\
      \Notice@String
    }%
  }%
}
\makeatother

%% file: sections/abstract.tex
Large language models (LLMs) act as tool-invoking agents, which introduces tool hallucination: selecting incorrect tools or generating invalid calls. We show that existing mitigation methods exhibit \textbf{Hallucination Escape}, a failure mode hidden by single-configuration evaluation: under controlled configuration changes, they reduce hallucination on the tuned configuration but increase it on others, canceling out the gain.
We find that hallucination spikes when a model's intrinsic tool-use tendencies conflict with the tool configuration, and that existing methods reinforce these tendencies.
Building on these findings, we propose \textbf{\method{}}, an inference-time method that combines conflict-aware gating with configuration-derived attention enhancement to mitigate tool hallucination while preventing escape.
Across six benchmarks and multiple models, \method{} reduces tool-selection hallucination by 9.0 pp, lowers the cross-configuration mean by 23.6 pp, and reduces total errors by 86.3\% in paired-query evaluation.
We hope this work encourages evaluation beyond a single tool configuration toward more reliable agents.

\vspace{-0.6em}

%% file: sections/01_introduction.tex
\section{Introduction}\label{sec:1}
\vspace{-0.3em}

\begin{figure*}[t]
\centering
\includegraphics[width=0.95\textwidth]{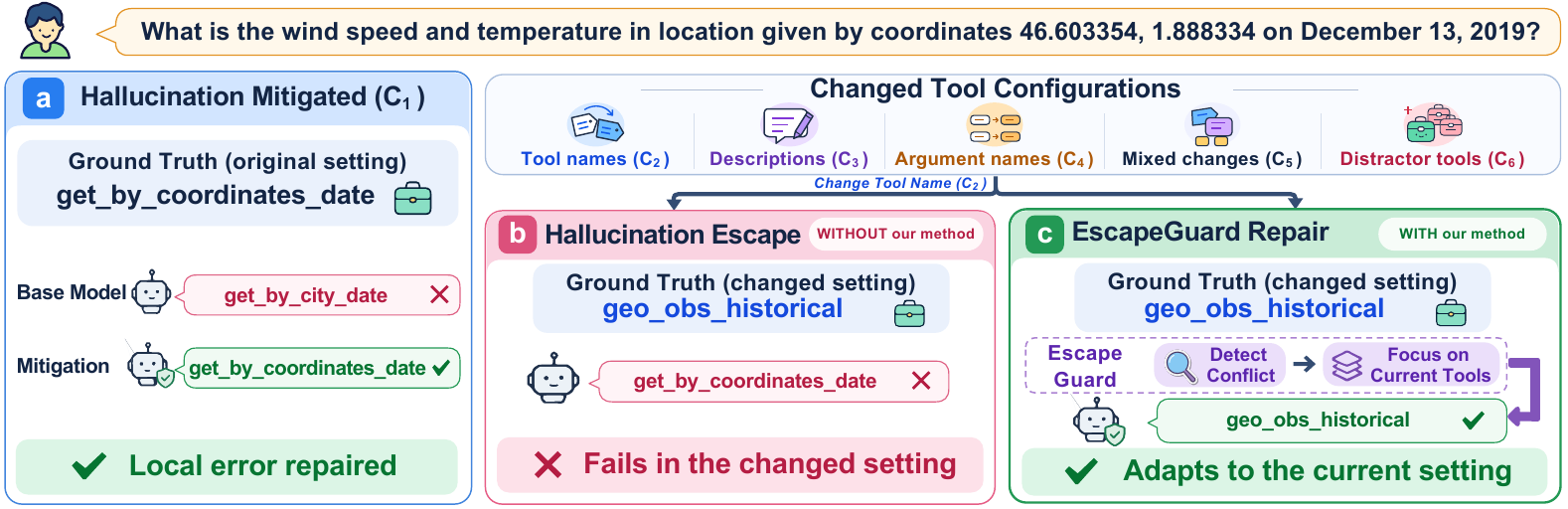}
\caption{\textbf{Hallucination escape and repair.}
(a)~Under the standard configuration~($C_1$), an existing mitigation method produces the correct call. (b)~After the tool is renamed~($C_2$), the same method still calls the old name.
(c)~\mbox{\method{}} adapts and produces the correct call.}
\label{fig:teaser}
\vspace{-0.5em}
\end{figure*}

Large language models (LLMs) increasingly serve as autonomous agents that call
external tools to complete tasks,
but this introduces tool hallucination: selecting
nonexistent or irrelevant tools (\emph{Tool-selection Hallucination}), or generating invalid arguments (\emph{Tool-usage
Hallucination})~\citep{xu2025,zhang2024,ke2026}. Unlike text
hallucination, which produces plausible but incorrect text, a hallucinated
tool call can trigger irreversible operations on real systems, making it a more consequential reliability
risk~\citep{zhang2024}. To address this problem, existing methods
fall into two families. \ding{182}~Training-based methods reshape model behavior
through modified data or objectives: Relign adds hesitation-based
deferral~\citep{xu2025}, Gorilla adopts retrieval-aware
fine-tuning~\citep{patil2024}, and PALADIN fine-tunes on fault-injected
trajectories~\citep{vuddanti2026}; \ding{183}~Steering-based methods instead
intervene on internal representations: LinSteer steers tool selection with
activation-difference vectors~\citep{wu2026}, and PRISMS applies
detector-gated directional interventions~\citep{ke2026}. Both families report substantial improvements on benchmarks (Figure~\ref{fig:teaser}a).

\textbf{Phenomenon (\S~\ref{sec:3}).} These improvements, however, are measured
only under the tool configuration each method is tuned on and ignore other configurations. We evaluate existing methods along two
dimensions: varying the tool configuration for a fixed query (Figure~\ref{fig:config}) and varying the
query under a fixed tool configuration (Table~\ref{tab:b1}). This evaluation exposes a failure mode that single-configuration evaluation does not reveal, which we term \mbox{\textbf{Hallucination Escape}}, where
mitigation reduces hallucination in the tuned setting but increases it in others
enough to outweigh the reduction (Figure~\ref{fig:teaser}b). Specifically,
\ding{182}~on the configuration side, all five methods reduce hallucination by
8.0~pp on average under their tuned configuration but increase it
by 13.5~pp across the remaining five, leaving no method with a lower
cross-configuration mean than the base model; \ding{183}~on the query side,
$\sim$40\% of repaired queries introduce new errors on paired
queries requiring a different tool. Tool-usage hallucination also shows limited sensitivity to all methods. These raise a question: what drives hallucination escape, and
why do existing methods exacerbate it?

\textbf{Investigation (\S~\ref{sec:4}).} To answer this, we examine whether
models hold intrinsic tool-use tendencies, since the behavior suggests
models disregard the runtime configuration. Concretely, we elicit outputs without the tool configuration and find that models
favor specific tools and argument structures
($\rho_{\mathrm{sel}}{=}0.65$), with hallucination rising sharply when they
conflict with the current configuration. We further validate this internally with layerwise probes, which
detect the conflict with 86.4\% accuracy and predict tool hallucination with an
AUROC of 0.82. Finally, existing methods reinforce
these tendencies (prior strength $+0.08$ on average), and stronger
reinforcement tracks larger cross-configuration error (Spearman
$\rho{=}0.96$), driving hallucination escape.

\textbf{Solution (\S~\ref{sec:5}).} Building on these findings, we propose
\mbox{\textbf{\method{}}}, an inference-time method that mitigates tool
hallucination while preventing hallucination escape (Figure~\ref{fig:teaser}c).
The key idea is to redirect attention to the runtime
configuration when a tendency conflict is detected (Figure~\ref{fig:f3}). Specifically, \mbox{\method{}} reuses the frozen conflict probes to flag high-conflict inputs and, only for them, strengthens
attention to tool names and descriptions during selection and to the selected
tool's schema during argument generation. This design ties the corrective signal
to the current configuration rather than a fixed tendency, so it adapts
as tools or queries change. Across six benchmarks on various models, \mbox{\method{}} reduces tool-selection
hallucination by 9.0~pp and argument-name hallucination by up to 67\% relative,
while suppressing escape along both dimensions: the cross-configuration mean
drops by 23.6~pp, and total errors fall by 86.3\% in paired-query evaluation.

In summary, our contributions are as follows:
\ding{182}~we identify and formally define hallucination escape, a
failure mode of existing mitigation methods that single-configuration evaluation hides, in which mitigation reduces hallucination
in the tuned configuration but increases it in others (\S~\ref{sec:3});
\ding{183}~we show that models hold intrinsic tool-use
tendencies that conflict with changed configurations, and that existing methods
reinforce these tendencies, contributing to escape (\S~\ref{sec:4});
\ding{184}~we propose \mbox{\method{}}, an inference-time method
that mitigates tool hallucination while effectively preventing hallucination escape across models (\S~\ref{sec:5}).

%% file: sections/02_related.tex
\section{Background and Related Work}
\label{sec:2}

\textbf{LLM-based Agents and Tool Use.} LLMs increasingly
act as agents that invoke tools.
Toolformer introduces self-supervised tool-call learning \citep{schick2023},
ReAct interleaves reasoning with action \citep{yao2023}, and ToolLLM scales
to real-world APIs \citep{qin2024}. As these capabilities expand,
so does the risk of erroneous calls.

\textbf{Tool Hallucination and Mitigation.} Tool hallucination occurs when a
tool call is inconsistent with the available tools or task \citep{xu2025,patil2024, liu2026,xiong2025}, including \textbf{tool-selection} hallucination
(invoking nonexistent or irrelevant tools) and \textbf{tool-usage} hallucination
(generating undefined parameters or ungrounded values). Mitigations include: \ding{182}~Training-based methods
reshape behavior through modified objectives: Relign \citep{xu2025}
adds hesitation-based deferral, PALADIN \citep{vuddanti2026} improves
fault recovery via annotated trajectories, and Gorilla \citep{patil2024}
reduces API hallucination via retrieval-aware tuning, and
RoTTuning~\citep{ye2024rotbench} trains on perturbed tool environments, while
RoTBench and AgentRobustBench~\citep{zhou2026robustbench} benchmark base-model
robustness; \ding{183}~Steering-based
methods intervene on internal activations: LinSteer \citep{wu2026}
steers tool selection via activations, and PRISMS
\citep{ke2026} applies detector-gated interventions per failure mode. Both evaluate only under the tuned-on configuration; we next test whether these gains generalize.

%% file: sections/03_escape.tex
\section{Phenomenon: Hallucination Escape}\label{sec:3}

Existing methods report reductions, yet their evaluations
keep tool definitions fixed, varying only the number, combination, and
availability of tools. In practice, tools with the same function differ in naming and arguments. We evaluate them under
varied tool configurations and queries, revealing a consistent failure mode we term hallucination escape.

\subsection{Evaluation Setup}
\label{sec:3.1}

\begin{figure*}[t]
\centering
\includegraphics[width=0.92\textwidth]{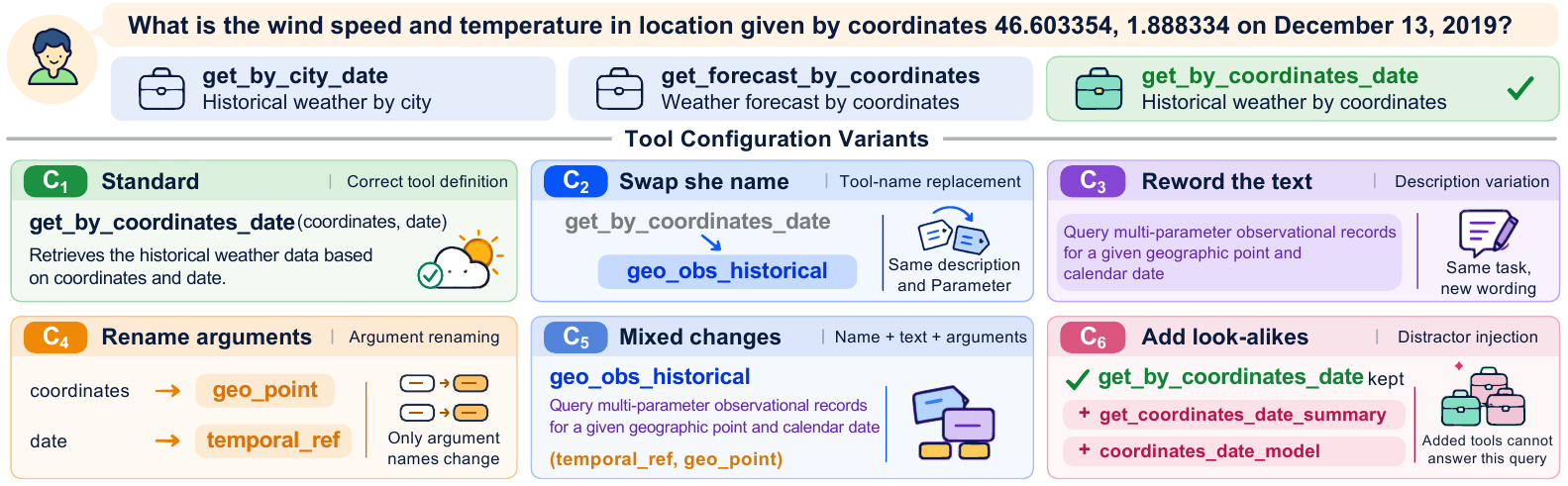}
\caption{\textbf{Tool configuration changes (C$_1$--C$_6$) illustrated with a weather API example.} Each variant modifies one or more aspects of the correct tool while ensuring a valid solution remains available.}
\label{fig:config}
\vspace{-0.8em}
\end{figure*}

\textbf{Evaluation Protocol.} The cross-configuration evaluation covers two aspects. \ding{182}~Tool Configuration: for each query, the tool list always contains one correct tool whose definition varies systematically. We construct six configurations (Figure~\ref{fig:config}): C$_1$ (standard) uses original names, descriptions, and arguments; C$_2$ (tool replacement) substitutes a functionally equivalent tool with a different name; C$_3$ (description change) rewrites functional descriptions; C$_4$ (argument renaming) replaces argument names; C$_5$ (mixed) simultaneously changes the tool name, description, and argument names; C$_6$ (distractor injection) adds tools with similar names but mismatched functionality.
All configurations were constructed using GPT-5.6-Terra, manually reviewed, and
validated to preserve a valid solution (Appendix~\ref{app:a}).
\ding{183}~Query Task: under the same tool list, we construct pairs of related
queries requiring different tools to test whether fixing one query introduces
errors on another (Appendix~\ref{app:b}).

\textbf{Methods, Models, and Data.}
We reproduce five methods spanning both categories.
For training-based methods, we adopt Relign~\citep{xu2025},
PALADIN~\citep{vuddanti2026}, and Gorilla~\citep{patil2024}; for
steering-based methods, we adopt LinSteer~\citep{wu2026} and
PRISMS~\citep{ke2026}. All use the original code and
are trained or calibrated on standard (C$_1$) data. We
use LLaMA-3.1-8B-Instruct \citep{meta2024llama} and Qwen-3.5-9B \citep{qwen2026}; base-model results for four recent
strong models are reported in Appendix~\ref{app:e.3}. Configuration data are sampled from BFCL~V3 and Seal-Tools ($\sim$1{,}000 examples
${\times}$ 6 configurations). For query changes, 120 pairs (240 queries) are
constructed from BFCL. All are repeated over three seeds;
see Appendix~\ref{app:c} for dataset selection.

\textbf{Evaluation Metrics.}
\ding{182}~For cross-configuration evaluation, we measure hallucination by: Tool-Selection Hallucination Rate (TSHR), the fraction of queries
where the model selects an unavailable or incorrect tool; and Tool-Usage Hallucination Rate (TUHR), conditional on correct
selection, decomposed into fabricated argument names (ANHR) and schema
violations or incorrect values (AVHR). Since C$_1$\,--\,C$_3$ and C$_6$ preserve the argument schema,
we report TSHR alone; for C$_4$ and C$_5$, which also modify arguments, we
report OHR~$= \text{TSHR} + (1 - \text{TSHR}) \times \text{TUHR}$.
Cross-configuration mean $\bar{H}(m)$ and variance $\mathrm{Var}(m)$ aggregate
these metrics. \ding{183}~For cross-query evaluation, we report Repair Rate (fraction of
errors corrected), Transfer Damage Rate (fraction of repairs introducing new
errors on the paired query), Conditional Net
Improvement (cNI)~$= \text{Repair} \times (1 - \text{Damage})$, and Net Error
Reduction (NER)~$=1-E_m/E_{\mathrm{base}}$ ($E$: total errors). Formal definitions in Appendix~\ref{app:d}.

\subsection{Evaluation Results}
\label{sec:3.2}
\begin{figure}[t]
\centering
\includegraphics[width=0.8\columnwidth]{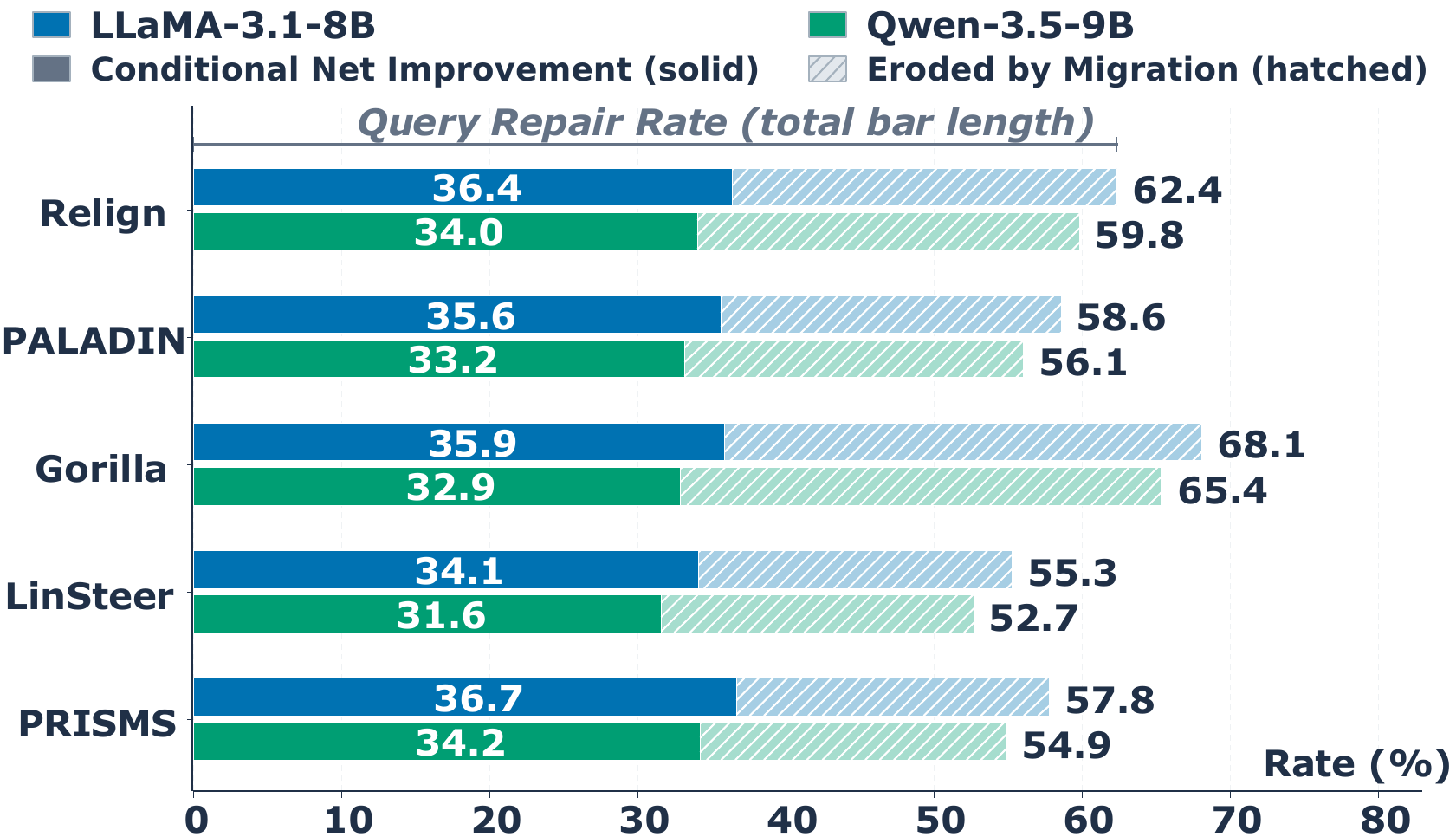}
\caption{\textbf{Cross-query hallucination escape (\%).}}
\label{fig:f2}
\vspace{-1em}
\end{figure}

We evaluate hallucination rates under six tool configurations
(Table~\ref{tab:m1}) and across paired queries that share a tool
list but require different tools (Figure~\ref{fig:f2}). We report two findings.

\begin{table*}[t]
\caption{{\textbf{Cross-configuration hallucination rates (\%).} Subscripts: change from the base (pp), computed from unrounded rates. $\bar{H}$: mean\,$\pm$\,std over three decoding seeds; 95\% paired-bootstrap CIs are in Table~\ref{tab:h17}.}}
\vspace{-0.2em}
\label{tab:m1}\centering\input{tables/m1}
\vspace{-0.5em}
\end{table*}

\textbf{\ul{Finding~1: Existing methods reduce the tuned tool-selection hallucination but worsen others.}} Both base models (Table 1) are highly sensitive to configuration changes: TSHR is
19.4\%\,--\,22.7\% under $C_1$ but surges to 47.5\%\,--\,50.4\% after tool-name
replacement ($C_2$). Existing methods amplify this imbalance: all reduce TSHR
on $C_1$ yet increase it by 2\,--\,9~pp on $C_2$, 1\,--\,6~pp on $C_3$ and
$C_6$, and 2\,--\,7~pp in OHR on $C_5$, while $C_4$ OHR changes remain marginal
(0.2\,--\,1.8~pp), suggesting OHR's usage component is largely
unaffected. Overall, no method ends with $\bar{H}$
below the base model, while $\mathrm{Var}$ rises by 66.0\% on average. The same
pattern holds across queries (Figure~\ref{fig:f2}): methods repair
target-query errors but with 37\%\,--\,50\% probability
introduce new errors on paired queries, limiting net improvement.
Four recent strong models are similarly sensitive (TSHR
3.9\%\,--\,8.6\% on $C_1$ vs.\ 19.6\%\,--\,31.4\% on $C_2$;
Appendix~\ref{app:e.3}).

\textbf{\ul{Finding~2: Tool-usage hallucination shows limited sensitivity to all methods.}}
To disentangle the two components of OHR, we compute TUHR
on correctly selected samples across all six configurations
(Appendix~\ref{app:e}). All methods change TUHR by at most 1.5~pp
on any configuration. Although base TUHR reaches 24\%\,--\,28\% on $C_4$ and $C_5$, existing
methods leave this component essentially untouched. The
marginal $C_4$ OHR reductions in Table~\ref{tab:m1} thus stem primarily from
small TSHR improvements, while the dominant TUHR component remains unchanged.
To control for denominator shifts from differing correct-selection sets,
Appendix~\ref{app:e.2} repeats the analysis on the shared subset, confirming
the same limited sensitivity (all changes ${\leq}\,0.8$~pp).

\subsection{Definition of Hallucination Escape}\label{sec:3.3}

The two findings above reveal a consistent pattern: existing methods shift
hallucination across configurations rather than reducing it overall. We
formalize this as follows.

\textbf{Definition 3.1 (Hallucination Escape).} Let $m$ be a mitigation method
and $\mathcal{C}=\{c_1,\ldots,c_K\}$ the configuration set. Method $m$
exhibits hallucination escape if: (1) $\exists\,\mathcal{C}_{\downarrow}
\subset \mathcal{C}$: $H(m, c_k) < H(\text{base}, c_k)$; (2)
$\exists\,\mathcal{C}_{\uparrow} \subset \mathcal{C} \setminus
\mathcal{C}_{\downarrow}$: $H(m, c_j) > H(\text{base}, c_j)$; (3) $\bar{H}(m)
\geq \bar{H}(\text{base})$.

All five methods satisfy this definition: they reduce hallucination on their tuned configuration yet
increase it on others, and no method significantly lowers the
cross-configuration mean: all methods except PRISMS significantly
raise it, and every method significantly increases the cross-configuration
variance (Appendix~\ref{app:h.6}).
The pattern holds across configurations and queries. Meanwhile,
tool-usage hallucination is largely immune to all methods, suggesting a
distinct mechanism. This raises a question: what
drives hallucination escape, and why do existing methods exacerbate it?

%% file: sections/04_mechanism.tex
\section{Investigation: The Mechanism of Hallucination Escape}
\label{sec:4}

The previous section shows existing mitigation methods reduce hallucination on their tuned
configuration but increase it on others. To understand what drives this, we
examine whether models hold intrinsic tool-use tendencies, as the observed
behavior suggests models disregard the runtime tool configuration. We first
test whether such tendencies exist and correlate with hallucination
(\S~\ref{sec:4.1}), then validate this finding internally through layerwise
probes (\S~\ref{sec:4.2}), and finally show how existing methods reinforce
these tendencies, contributing to hallucination escape
(\S~\ref{sec:4.3}).

\subsection{Intrinsic Tendencies and Hallucination}\label{sec:4.1}

We investigate two aspects: whether models hold
intrinsic tool-use tendencies that persist without access to the runtime
configuration, and how hallucination relates to conflicts between these
tendencies and the current configuration.

\textbf{Identifying Intrinsic Tool-Use Tendencies.}
We examine what models generate without the tool configuration: if
outputs are consistent across samples, the model holds a strong intrinsic
tendency. We evaluate six models on 300 queries randomly selected from BFCL~v3, Seal-Tools, and
Glaive Function Calling~V2 using closed-book elicitation ($K{=}10$ samples per
query $q$, temperature 0.7; Appendix~\ref{app:c.2}).
For tool selection, we remove
the tool list and measure the prior strength as the frequency of the most
commonly generated tool $t$:
{\small $S_{\mathrm{prior}}^{\mathrm{sel}}(q) = \max_{t} \frac{1}{K}
\sum_{j=1}^{K} \mathbf{1}[t_j = t]$}.
For tool usage, we provide the tool name but omit its schema and compute
{\small $S_{\mathrm{prior}}^{\mathrm{use}}(q) = \frac{1}{K}
\sum_{i=1}^{K} \operatorname{Jaccard}(P_i, P_{\mathrm{mode}})$},
where $P_i$ is the argument-name set in sample $i$ and $P_{\mathrm{mode}}$ the
most frequent set. Across models, mean selection priors range from 0.61 to 0.78
and usage priors from 0.57 to 0.71 (Table~\ref{tab:f1}), indicating that models  favor specific tools
and argument structures even without the runtime configuration.

\begin{figure*}[t]
\centering
\includegraphics[width=0.8\textwidth]{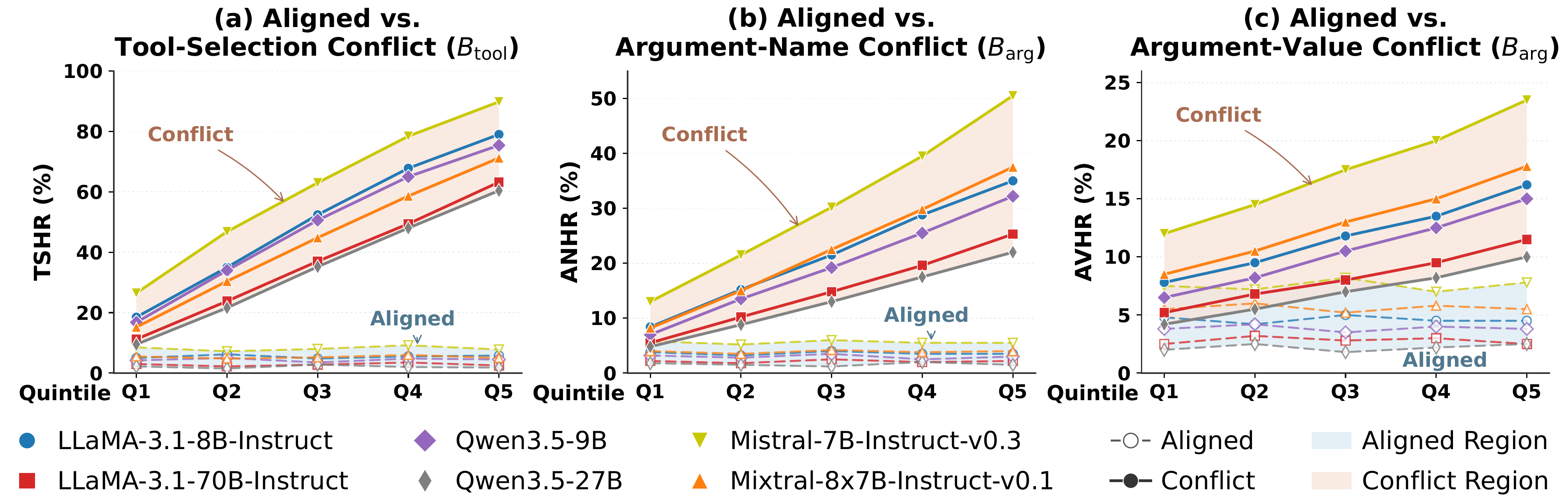}
\caption{\textbf{Hallucination rate by prior-strength quintile under aligned and conflict settings.}
(a)~tool-selection conflict under $B_{\mathrm{tool}}$;
(b)~argument-name and (c)~value conflict under
$B_{\mathrm{arg}}$. Q1--Q5: weakest to strongest tendency.}
\label{fig:f4}
\end{figure*}

\textbf{Tendency Conflict and Hallucination.}
We next test whether these tendencies become harmful when they disagree with
the current configuration. For each query we construct three matched settings:
\emph{aligned}~(A), where the correct tool and schema match the elicited
tendency; \emph{tool-conflict}~($B_{\mathrm{tool}}$), which replaces the
preferred tool with a functionally equivalent alternative under a different
name and rewritten description; and
\emph{argument-conflict}~($B_{\mathrm{arg}}$), which keeps the tool name but
modifies argument names or types. Both conflict settings preserve valid
solutions. We compare TSHR between A and $B_{\mathrm{tool}}$
(Figure~\ref{fig:f4}a), and TUHR between A and $B_{\mathrm{arg}}$
(Figure~\ref{fig:f4}b,c). To examine how prior strength modulates
hallucination, we rank queries by prior strength and divide them into five
equal-sized quintiles ($Q_1$\,--\,$Q_5$, weakest to strongest), measuring
hallucination within each under both conditions. As shown in Figure~\ref{fig:f4}, under alignment, hallucination remains
uniformly low regardless of prior strength, whereas under conflict it
increases monotonically from $Q_1$ to $Q_5$. This correlation is strongest for tool selection
($\rho{=}0.58$\,--\,$0.72$) and argument name
($\rho{=}0.50$\,--\,$0.64$), while argument value shows only a modest effect
($\rho{=}0.30$\,--\,$0.38$). Per-model breakdowns in
Appendix~\ref{app:f}. These suggest that intrinsic tendencies are
benign when aligned with the configuration but are associated with
substantially increased hallucination when a conflict arises, with severity
scaling with tendency.

The empirical experiments reveal that models exhibit stable intrinsic tendencies
at both the selection and usage levels. When these tendencies align with the
runtime configuration, hallucination remains uniformly low; when they conflict,
hallucination rises sharply and scales with tendency. This evidence is behavioral, so we next seek validation in the model's internal representations.

\subsection{Internal Evidence of Tendency Conflict}\label{sec:4.2}
To move beyond behavioral evidence, we probe whether the conflict between
intrinsic tendencies and the runtime configuration is separable in the model's
representations and whether this signal is associated with tool
hallucination.

\textbf{Layerwise Conflict Probing.}
We probe LLaMA-3.1-8B and Qwen-3.5-9B on 1,200 queries (the 300 from
\S~\ref{sec:4.1} plus 900 more). We extract the hidden state
$\mathbf{h}^{(l)} \in \mathbb{R}^{d}$ at the last input token from every
decoder layer and train a logistic-regression probe at each layer
(Appendix~\ref{app:g.2}):
\begin{equation}
    p_r^{(l)}(x) = \sigma\!\left(
    (\mathbf{w}_r^{(l)})^\top \mathbf{h}^{(l)}(x) + b_r^{(l)}
    \right), \quad r \in \{\mathrm{sel}, \mathrm{arg}\},
\end{equation}
where $x$ denotes the serialized input, $\mathbf{w}_r^{(l)}$ and $b_r^{(l)}$
are learned probe parameters, and $r$ indexes the tool-selection and
argument-generation stages, respectively. As shown in Table~\ref{tab:m2}, probes achieve 85\%\,--\,87\% accuracy for tool selection and 77\%\,--\,79\% for argument name,  and control experiments rule out potential confounds (Appendix~\ref{app:g.5}\&\ref{app:g.5b}). Extended results across all backbones in Appendix~\ref{app:g.3}.

\begin{table}[h]
    \vspace{-0.5em}
    \centering
    \caption{\textbf{Conflict detection and hallucination prediction.}}
    \vspace{-0.5em}
    \label{tab:m2}
    \input{tables/m2.tex}
    \vspace{-0.5em}
\end{table}

\textbf{Predicting hallucination from conflict.}
Distinguishing A from B alone does not ensure the detected signal relates to
hallucination. We thus test whether it predicts hallucination
\emph{within} conflicting inputs. We use the peak-layer probe output as a
conflict score
{\small $P_{\mathrm{conflict}}^r(x) = p_r^{(l_r^*)}(x)$},
divide conflicting inputs into quintiles, and measure hallucination rates
per quintile. For both models (Table 2), TSHR and ANHR
increase monotonically from $Q_1$ to $Q_5$, the extreme conflict-score quintiles (e.g., LLaMA-3.1-8B: TSHR
19.6\%\,$\to$\,81.2\%, ANHR 8.4\%\,$\to$\,35.6\%), with AUROC
0.81\,--\,0.84 (selection) and 0.74\,--\,0.75 (argument name).

These results suggest that tendency conflict is not only behavioral but also
encoded internally, with stronger conflict associated with higher hallucination.
Armed with these probes, we next examine how existing methods affect these
tendencies and their link to hallucination escape.

\begin{figure*}[t]
\centering
\includegraphics[width=1.0\textwidth]{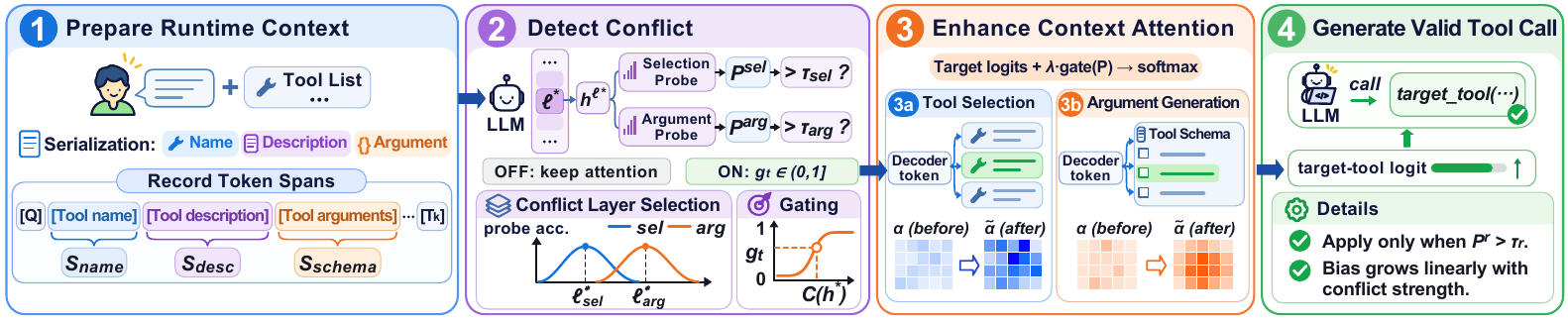}
\caption{\textbf{Overview of \method{}.} Given a query and tool list,
\method{} records tool-related token spans and extracts frozen-LLM hidden
states (\ding{182}). Conflict probes score each input per stage and
intervene only above a threshold (\ding{183}). When activated,
attention is enhanced toward tool names and descriptions during selection, or
the selected tool's schema during argument generation (\ding{184}).}
\label{fig:f3}
\end{figure*}

\subsection{Hallucination Escape in Existing Methods}\label{sec:4.3}

The previous sections establish that hallucination rises when intrinsic
tendencies conflict with the runtime configuration. A natural hypothesis is
that existing methods reinforce these tendencies, explaining why they improve
aligned configurations at the cost of conflicting ones. We test this by
measuring how each method changes tendency strength.

\textbf{Setup.}
We repeat closed-book elicitation before and after applying each method,
computing the prior-strength change only for queries whose preferred tool
remains unchanged:
{\small $\Delta S_m = \frac{1}{|\mathcal{Q}_m^{\mathrm{ret}}|}
\sum_{q \in \mathcal{Q}_m^{\mathrm{ret}}}
\bigl[S_m(q) - S_{\mathrm{base}}(q)\bigr]$},
where $\mathcal{Q}_m^{\mathrm{ret}}$ denotes these queries. To quantify the severity of escape, we define the weighted
Escape Ratio $E_w(m) = O(m)/I(m)$, where $I(m)$ and $O(m)$ aggregate total
improvement and deterioration across configurations; values above~1 indicate
total deterioration exceeding improvement.

\begin{table}[h]
    \vspace{-0.5em}
    \centering
    \caption{\textbf{Tendency reinforcement and escape.}}
    \label{tab:m3}
    \vspace{-0.5em}
    \input{tables/m3.tex}
    \vspace{-0.5em}
\end{table}

\textbf{Results.}
As shown in Table~\ref{tab:m3}, 76\%\,--\,93\% of queries retain their
preferred tool after mitigation (denoted Ret. in Table~\ref{tab:m3}), with
prior strength increasing by 0.04\,--\,0.14 across all methods. Stronger
reinforcement goes with a larger increase in $\bar{H}$ (Spearman $\rho{=}0.96$
over ten method--model pairs, $p<0.001$); $E_w$ shows the same trend
($\rho{=}0.94$). The same holds across
queries: reinforced preferences overgeneralize to similar queries needing
other tools. Removing the prior-aligned part of LinSteer's steering
vector removes escape yet keeps most of the $C_1$ gain
(Appendix~\ref{app:f.decomp}).

Together, \S\ref{sec:4.1}\,--\,\S\ref{sec:4.3} provide
a coherent picture: models hold intrinsic tool-use tendencies that are
associated with increased hallucination under conflict, are encoded in internal
representations, and are reinforced by existing methods. This motivates a
conflict-aware approach that mitigates tool hallucination while preventing escape, introduced next.

%% file: sections/05_method.tex
\section{Solution: EscapeGuard}\label{sec:5}

The analysis traces hallucination escape to intrinsic
tendencies that existing methods reinforce. This motivates \textbf{\method{}}, an inference-time method that
detects tendency conflicts and redirects attention to the runtime settings
(Figure~\ref{fig:f3}; design in \S\ref{sec:5.1}, evaluation in \S\ref{sec:5.2}\&\S\ref{sec:5.3}).

\begin{table*}[t]
\vspace{-0.5em}
\caption{\textbf{TSHR ($\downarrow$\%) across six backbones and six benchmarks.}
Subscripts: change from Base.}
\vspace{-0.5em}
\label{tab:m4}\centering\input{tables/m4}
\vspace{-0.5em}
\end{table*}

\begin{table}[t]
    \centering
    \caption{\textbf{Usage-level macro averages (\%).}}
    \vspace{-0.5em}
    \label{tab:m5}
    \input{tables/m5}
    \vspace{-0.5em}
\end{table}

\subsection{Design of \method{}}\label{sec:5.1}

The core idea of \method{} is to detect when a model's intrinsic tendencies
conflict with the runtime tool configuration and, only in those cases,
strengthen the model's attention to the relevant tool definitions. This
requires addressing three questions: \textbf{what} to attend to, \textbf{when} to
intervene, and \textbf{how} to enhance attention. Accordingly, \method{} operates
in three stages: input preparation, conflict-aware gating, and context-specific
attention enhancement.

\textbf{Stage \ding{182}: Input Preparation.}
Given a query $q$ and tool configuration $C$, \method{} first identifies the
token spans that serve as attention targets in later stages. For each tool $t \in C$,
it records the token spans of its name ($S_{\mathrm{name}}(t)$),
description ($S_{\mathrm{desc}}(t)$), and argument schema
($S_{\mathrm{schema}}(t)$) during input serialization. These spans define
\textbf{what} the model should attend to when intervention is triggered.

\textbf{Stage \ding{183}: Conflict-Aware Gating.} \method{} next decides \textbf{when} to intervene.
It reuses the frozen probes of \S\ref{sec:4.2} to score conflict for each stage $r \in \{\mathrm{sel}, \mathrm{arg}\}$:
\begin{equation}
P_{\mathrm{conflict}}^{r}(q,C) = \sigma\!\left(
\mathbf{w}_{r}^{\top}\mathbf{h}^{(l_r^*)}(q,C) + b_r
\right),
\end{equation}
where $l_r^*$ is the representative conflict layer. Intervention activates
only when $P_{\mathrm{conflict}}^{r}(q,C) > \tau_r$; otherwise the original
computation is preserved. This gating mechanism ensures that \method{} modifies
only high-conflict inputs, leaving low-conflict ones untouched.

\textbf{Stage \ding{184}: Context-Specific Attention Enhancement.}
When the gate activates, \method{} enhances attention toward the current tool
configuration, addressing \textbf{how} to intervene. During tool selection, the
target span covers all tool names and descriptions:
$S_{\mathrm{sel}}(C) = \bigcup_{t \in C}(S_{\mathrm{name}}(t) \cup
S_{\mathrm{desc}}(t))$. After the model generates a tool $t$, the target
shifts to its schema span $S_{\mathrm{arg}}(C,t) = S_{\mathrm{schema}}(t)$.
For an attention head at the intervention layer, the logit
$z_{ij} = \mathbf{q}_i^\top \mathbf{k}_j / \sqrt{d}$ from decoding token $i$
to context token $j$ is modified as
\begin{equation}
\begin{aligned}
\tilde{z}_{ij} = z_{ij} &+ \lambda_r \frac{P_{\mathrm{conflict}}^{r}(q,C)-\tau_r}{1-\tau_r}\\
&\times \mathbf{1}\!\left[P_{\mathrm{conflict}}^{r}(q,C) > \tau_r\right]
\mathbf{1}\!\left[j \in S_r\right],
\end{aligned}
\end{equation}
where $\lambda_r$ is the maximum intervention strength, $\tau_r$ is the
conflict threshold, and $S_r$ denotes $S_{\mathrm{sel}}$ or $S_{\mathrm{arg}}$. The factor $(P_{\mathrm{conflict}}^r - \tau_r)/(1-\tau_r)$
rescales the above-threshold conflict score to $[0,1]$, so the bias grows
linearly with conflict strength rather than switching on abruptly.
Attention is then computed via
$\tilde{\alpha}_{ij} = \mathrm{softmax}_j(\tilde{z}_{ij})$.

All three stages derive from the current tool configuration rather
than a fixed correction, allowing \method{} to adapt to tool
renaming, replacement, and schema changes.  It
mitigates tool hallucination by redirecting attention to the runtime
configuration while avoiding the tendency reinforcement behind
hallucination escape in existing methods.

\subsection{Experimental Setup}\label{sec:5.2}

We now evaluate \method{} across diverse
models, benchmarks, and tool configurations.

\textbf{Datasets.}
We use six benchmarks: BFCL~\citep{patil2025}, Seal-Tools~\citep{wu2024},
RelyToolBench~\citep{xu2025}, API-Bank~\citep{li2023},
MetaTool~\citep{huang2024}, and APIBench (subsets, pool sizes, and metric
applicability in Table~\ref{tab:h2}). All evaluation data is disjoint from
the probing and tuning data.

\textbf{Models, Baselines, and Metrics.}
We use the six backbones from \S\ref{sec:4}:
LLaMA-3.1-8B/70B-Instruct~\citep{meta2024llama},
Qwen-3.5-9B/27B~\citep{qwen2026},
Mistral-7B-Instruct-v0.3~\citep{mistral2024}, and
Mixtral-8x7B-Instruct-v0.1~\citep{mixtral2024}.
All six baselines, including
RoTTuning~\citep{ye2024rotbench} trained on perturbed tool configurations
under the same data and budget, are compared with \method{} on LLaMA-3.1-8B
and Qwen-3.5-9B; the other four backbones test cross-model generalization, and DeepSeek-V4.1-Flash
and GLM-5.3-Flash are evaluated in Appendix~\ref{app:h.12}.
We report TSHR, TUHR (split into ANHR and AVHR), and OHR (\S\ref{sec:3});
Table~\ref{tab:h2} lists per-benchmark metric applicability.

\textbf{Intervention Setup.}
We set conflict threshold $\tau_r{=}0.6$ and attention bias $\lambda_r{=}1.8$ on held-out data; both are
stable over a broad range (Appendix~\ref{app:h.8}). Intervention layers per
model, including the softmax-attention layers used for hybrid architectures,
are in Appendix~\ref{app:h.1}.

\begin{table*}[t]
\caption{\textbf{Comparison with existing methods on LLaMA-3.1-8B-Instruct.}
Lower is better. Bold\,=\,best.}
\vspace{-0.5em}
\label{tab:m6}\centering\input{tables/m6}
\end{table*}

\begin{table*}[t]
\caption{\textbf{RoTTuning and \method{} under C$_1$\,--\,C$_6$.} Subscripts: change from
the base model in Table~\ref{tab:m1}, computed from unrounded rates. $C_4$/$C_5$: OHR; others: TSHR.}
\vspace{-0.5em}
\label{tab:m7}\centering\input{tables/m7}
\end{table*}

\subsection{Experimental Results}\label{sec:5.3}

We evaluate \method{} from four perspectives: overall effectiveness,
baseline comparison, cross-configuration robustness, and
component ablation study.

\textbf{Overall Effectiveness.}
As shown in Table~\ref{tab:m4}, \method{} reduces TSHR for every
model-benchmark pair, with improvements ranging from 4.3\,pp to 14.7\,pp. At
the usage level (Table~\ref{tab:m5}; subscripts: change from Base), \method{} consistently lowers TUHR and
ANHR across all six models. The improvement in AVHR is substantially smaller,
consistent with the weaker association between argument-value hallucination and
tendency conflict observed in \S\ref{sec:4}. These gains come at modest cost:
inference overhead is 4\%\,--\,6\% of base latency (Appendix~\ref{app:h.9}).

\textbf{Baseline Comparison.}
We compare \method{} with all six baselines on LLaMA-3.1-8B using BFCL,
Seal-Tools, API-Bank, and APIBench. As shown in Table~\ref{tab:m6}, \method{}
achieves the lowest TSHR, TUHR, and ANHR across all four benchmarks, with more
stable improvements than existing methods. While baselines reduce TSHR by
3.0\,--\,5.3\,pp on average, \method{} achieves 9.0\,--\,12.5\,pp reductions. The
advantage is most pronounced for ANHR, where \method{} roughly halves the base
rate while existing methods show marginal changes. The same pattern holds on
Qwen-3.5-9B (Appendix~\ref{app:h.5}). RoTTuning reduces TSHR by
4\,--\,6\,pp, comparable to the other baselines, and ANHR by about 1\,pp.

\textbf{Cross-configuration Robustness.}
To directly test whether \method{} prevents hallucination escape, we evaluate
under the same $C_1$\,--\,$C_6$ configurations from \S\ref{sec:3}. As shown
in Table~\ref{tab:m7}, \method{} reduces hallucination across all
configurations. Under $C_1$\,--\,$C_3$, rates converge to 5.7\,--\,8.6\%,
effectively narrowing the performance gap among configurations. Mixed changes
($C_5$) see OHR reductions of 33.3\,--\,34.2\,pp. The smallest gains appear
under $C_6$ (13.2\,--\,13.7\,pp), plausibly because injected distractors
dilute the enhanced attention signal. For LLaMA and Qwen, the
cross-configuration mean decreases by 24.2 and 23.1\,pp while variance falls
by 41.8\% and 40.0\%. For paired-query evaluation (Table~\ref{tab:m9}),
\method{} achieves 87.8\%\,--\,90.4\% cNI and reduces total errors by
85.4\%\,--\,87.1\% (Table~\ref{tab:h18}), maintaining a high repair rate
while introducing substantially less transfer damage than baselines.

\textbf{Extended Evaluation.}
RoTTuning lowers $\bar{H}$ by 5.6/5.2\,pp but worsens $C_3$ and $C_6$ by
3.1\,--\,5.1\,pp (Table~\ref{tab:m7}), outside its training
perturbations. On DeepSeek-V4.1-Flash and GLM-5.3-Flash, \method{} lowers
$\bar{H}$ by 11.6/10.1\,pp, whereas LinSteer and PRISMS leave it above Base
(Table~\ref{tab:h15}). Under interface changes, \method{} lowers OHR from
48.4\%/44.6\% to 28.3\%/25.5\%, with fewer selection, naming, and
schema errors (Appendix~\ref{app:h.14}).

\begin{table}[h]
    \centering
    \caption{\textbf{Paired-query performance (\%).} NER with 95\% CI.}
    \vspace{-0.5em}
    \label{tab:m9}
    \input{tables/m9}

\end{table}

\textbf{Ablation Study.}
We ablate four components on LLaMA-3.1-8B and Qwen-3.5-9B
(Appendix~\ref{app:h.10}, Table~\ref{tab:m10}). Each contributes: the
conflict gate avoids unnecessary reweighting; stage-specific enhancement
outperforms random context token enhancement; the correct conflict layers are
essential; and input-dependent enhancement outperforms fixed steering vectors.

Overall, among all evaluated methods, only \method{} reduces hallucination on
every configuration while achieving the lowest TSHR, TUHR, and ANHR against all
six baselines and cutting paired-query errors by over 85\%. These results show
that conflict-aware, configuration-grounded intervention mitigates tool
hallucination without the escape that undermines existing approaches.

%% file: sections/06_discussion.tex
\vspace{-0.2em}
\section{Discussion}\label{sec:6}

Since $C_2$\,--\,$C_6$ are controlled surface-form changes, escape in
multi-turn settings, dynamic registries, and larger tool libraries remains to be
measured, although \method{} extends to missing tools and interface changes
except for value conversion (Appendices~\ref{app:h.13} and~\ref{app:h.14}).
Gains are smaller on argument-value errors, which tendency conflict explains
only weakly, and under distractor injection ($C_6$), where confusable tools
likely dilute the enhanced attention. Addressing these cases and extending to
broader architectures are promising future directions.

%% file: sections/07_conclusion.tex
\section{Conclusion}\label{sec:7}

\vspace{-0.2em}
We identify hallucination escape, where mitigation methods reduce tool
hallucination on their tuned configuration but increase it on others, and trace
it to intrinsic tool-use tendencies that existing methods reinforce. Our
\method{} detects tendency conflicts and redirects attention to
the runtime configuration, reducing tool-selection hallucination by 9.0\,pp
while suppressing escape. Its gains hold across six backbones and six benchmarks without changing model weights. We hope this work encourages evaluation beyond a
single tool configuration toward more reliable tool-using agents whose tool interfaces keep evolving.

%% file: appendices/A.tex
\section{Tool Configuration Construction}\label{app:a}

\subsection{Configuration Construction Rules}
\label{app:a.1}

Every tool list in $C_1$\,--\,$C_6$ contains the correct tool and holds at most 10 tools, so that differences across configurations are not confounded by context length.

\textbf{Tool Replacement Rules (\texorpdfstring{$C_2$}{C2}). }The replacement tool name uses different lexical forms to reduce direct matching to the original. For example, \code{currency_conversion.get_rate} \(\rightarrow\) \code{fx_historical_lookup} (rather than \code{get_currency_rate}, which retains the keywords ``currency'' and ``rate,'' allowing the model to infer functionality). The functional description remains complete and accurate, so the correct answer remains obtainable from the description.

\textbf{Description Change Rules (\texorpdfstring{$C_3$}{C3}). }The tool name remains unchanged while the functional description's wording and emphasis are altered, reflecting the versatility of the same tool across different scenarios. For example, the description of \code{currency_conversion.get_rate} is ``Get the currency conversion rate from one currency to another at a specified date,'' which is changed to ``Retrieve the foreign exchange mid-market reference between two ISO 4217 currency codes for a given valuation date.'' The correct tool-query mapping remains intact.

\textbf{Argument Renaming Rules (\texorpdfstring{$C_4$}{C4}). }Argument names are replaced with alternative surface forms to reduce reliance on familiar names. For example, \code{from_currency} \(\rightarrow\) \code{source_unit}, \code{to_currency} \(\rightarrow\) \code{target_unit}, \code{date} \(\rightarrow\) \code{temporal_ref}. The functional descriptions of arguments remain accurate, while their surface forms change.

\textbf{Mixed Changes Rules (\texorpdfstring{$C_5$}{C5}). }The tool name, functional description, and argument names are replaced simultaneously, with argument order randomly shuffled. This is the most comprehensive scenario and the closest to real-world deployments.

\textbf{Distractor Tool Construction Rules (\texorpdfstring{$C_6$}{C6}). }Injected distractor tools satisfy the following conditions: (1) the tool name is highly similar to high-frequency tool names in the model's prior (e.g., \code{currency_conversion.get_rate_forecast} vs.~\code{currency_conversion.get_rate}), maximizing naming confusion; (2) the description partially overlaps with but does not fully match the correct tool (e.g., ``Get the forecasted future conversion rate trend between two currencies'' vs.~``Get the currency conversion rate from one currency to another at a specified date''), testing the model's fine-grained discrimination; (3) 2--3 distractor tools are injected per query, ensuring the total tool list length does not exceed 10 (to avoid confounding factors from excessive context length).

\textbf{Configuration Style Note. }The five mitigation methods analyzed in \S\ref{sec:3} are trained/calibrated under the standard naming style \(C_1\), the default in each original paper, so \(C_2\)--\(C_6\) constitute out-of-distribution test scenarios for them. RoTTuning, added as a baseline in \S\ref{sec:5}, is instead trained on perturbed tool and argument names following its original paper.

\subsection{Configuration Construction Examples}\label{app:a.7}

The following five examples are drawn from the BFCL V3 single-turn portion, Glaive Function Calling V2, and Seal-Tools, covering five domains: finance, communication, translation, meteorology, and travel, to illustrate the systematic changes in tool configuration under \(C_1\)--\(C_6\) for the same query.

\Needspace{10\baselineskip}
\subsubsection*{Example 1: Exchange Rate Query}

\textbf{Query.} ``Find the conversion rate from Euro to Dollar at January 1, 2022''

\textbf{Original tool list} (3 tools; correct tool is \code{currency_conversion.get_rate}):

\input{tables/schema1.tex}

\(C_1\)--\(C_6\) variants (only the correct tool \code{currency_conversion.get_rate} changes; other tools are unchanged):

\def\ATabCaption{\textbf{Configuration for example 1. Only the correct tool changes in \(C_1\)--\(C_5\); \(C_6\) adds distractors.}}\def\ATabLabel{tab:a1}\par\addvspace{\BlockSep}\input{tables/a1.tex}\par\addvspace{\BlockSep}

\Needspace{10\baselineskip}
\subsubsection*{Example 2: Email Sending}

\textbf{Query.} ``I need to send an email to my boss. The recipient's email is boss@company.com. The subject is `Project Update'. And the message is `Dear Boss, I have completed the project as per the given deadline.'\,''

\textbf{Original tool list} (2 tools; correct tool is \code{send_email}):

\input{tables/schema2.tex}

\(C_1\)--\(C_6\) variants (only the correct tool \code{send_email} changes; other tools are unchanged):

\def\ATabCaption{\textbf{Configuration for example 2. Only the correct tool changes in \(C_1\)--\(C_5\); \(C_6\) adds distractors.}}\def\ATabLabel{tab:a2}\par\addvspace{\BlockSep}\input{tables/a2.tex}\par\addvspace{\BlockSep}

\Needspace{10\baselineskip}
\subsubsection*{Example 3: Translation}

\textbf{Query.} `Tell me how to say ``Today is a beautiful day'' in Amharic.'

\textbf{Original tool list} (5 tools; correct tool is \code{translateAmharic}):

\input{tables/schema3.tex}

\(C_1\)--\(C_6\) variants (only the correct tool \code{translateAmharic} changes; other tools are unchanged):

\def\ATabCaption{\textbf{Configuration for example 3. Only the correct tool changes in \(C_1\)--\(C_5\); \(C_6\) adds distractors.}}\def\ATabLabel{tab:a3}\par\addvspace{\BlockSep}\input{tables/a3.tex}\par\addvspace{\BlockSep}

\Needspace{10\baselineskip}
\subsubsection*{Example 4: Meteorological Data}

\textbf{Query.} ``What is the wind speed and temperature in location given by coordinates 46.603354, 1.888334 on December 13, 2019?''

\textbf{Original tool list} (3 tools; correct tool is \code{weather.get_by_coordinates_date}):

\input{tables/schema4.tex}

\(C_1\)--\(C_6\) variants (only the correct tool \code{weather.get_by_coordinates_date} changes; other tools are unchanged):

\def\ATabCaption{\textbf{Configuration for example 4. Only the correct tool changes in \(C_1\)--\(C_5\); \(C_6\) adds distractors.}}\def\ATabLabel{tab:a4}\par\addvspace{\BlockSep}\input{tables/a4.tex}\par\addvspace{\BlockSep}

\Needspace{10\baselineskip}
\subsubsection*{Example 5: Flight Booking}

{\sloppy\looseness=-1 \textbf{Query.} ``I need to book a flight to Paris. I want to leave on the 15th of next month and return on the 30th. We are a group of 5.''\par}

\textbf{Original tool list} (2 tools; correct tool is \code{book_flight}):

\input{tables/schema5.tex}

\(C_1\)--\(C_6\) variants (only the correct tool \code{book_flight} changes; other tools are unchanged):

\def\ATabCaption{\textbf{Configuration for example 5. Only the correct tool changes in \(C_1\)--\(C_5\); \(C_6\) adds distractors.}}\def\ATabLabel{tab:a5}\par\addvspace{\BlockSep}\input{tables/a5.tex}\par\addvspace{\BlockSep}

\subsection{Configuration Quality Control}\label{app:a.8}

All configurations ($C_2$\,--\,$C_6$) were initially generated with LLM assistance and then independently verified and refined by two authors for consistency. The construction process followed three stages.

\textbf{LLM-assisted generation and independent verification.\enspace}Configuration variants were initially generated by prompting an LLM with the rules in Section~\ref{app:a.1}. The full set of variants was then independently reviewed and refined through manual verification, correcting cases where the generated output violated a construction rule or introduced unintended cues. Reviewers worked from the same generated candidates without sharing intermediate edits.

\textbf{Reconciliation.\enspace}The two sets were compared at the tool-name, description, argument-name, and distractor levels. Discrepancies (averaging 8.3\% of entries across configurations) were resolved by discussion, preferring for the variant that more completely satisfied the stated construction rules.

\textbf{Automated validation.\enspace}After reconciliation, automated checks verified that (1)~every configuration variant preserves a valid reference solution (the correct tool and its required arguments remain present and invocable), (2)~$C_2$ tool names share no content words with the original, (3)~$C_3$ descriptions preserve the original tool's functional scope, (4)~$C_4$ argument names differ from the originals in surface form while the schema remains structurally valid, and (5)~$C_6$ distractor tools do not duplicate any existing tool in the list. All queries passed these checks; the validation scripts will be released with the code.

%% file: appendices/B.tex
\section{Query Pair Construction Rules}\label{app:b}

Query pairs satisfy: (1) they share the same tool list and each has a unique correct tool; (2) they are semantically related but involve different tasks; (3) the correct tools' functional descriptions are sufficient to distinguish them. Pairs come from the BFCL V3 single-turn multi-tool portion (30 pairs in each of four functional domains); Table~\ref{tab:b1} lists examples.

\vspace{-0.5em}
\begin{table}[!htbp]\caption{\textbf{Query pair examples from the BFCL V3 multiple subset.}}
\label{tab:b1}\centering\input{tables/b1.tex}\end{table}
\vspace{-0.5em}

%% file: appendices/C.tex
\section{Dataset Selection and Data Construction}
\label{app:c}

This appendix describes the data used for controlled evaluation
in \S~\ref{sec:3}, behavioral analysis in
\S~\ref{sec:4.1}, and internal representation analysis in
\S~\ref{sec:4.2}.

\subsection{Controlled and Paired-Query Evaluation}
\label{app:c.1}

The controlled evaluation uses the single-turn portions of BFCL V3 and Seal-Tools, whose explicit tool definitions and reference calls support controlled configuration changes while preserving a valid solution. About 1,000 source queries yield about 6,000 evaluations across $C_1$--$C_6$, each run with three random seeds to account for sampling variance.

Paired-query evaluation uses 120 BFCL V3 single-turn multi-tool pairs (240 queries; 30 in each of four domains). Each pair shares a tool list but requires different correct tools (Appendix~\ref{app:b}), isolating query-level effects.

\subsection{Behavioral Analysis Set}
\label{app:c.2}
\label{app:f.1}

\S~\ref{sec:4.1} uses a behavioral set of 300 queries sampled equally from BFCL, Seal-Tools, and Glaive Function Calling V2 (100 each), which reduces dataset-specific bias while retaining diverse tool names, descriptions, and argument schemas.

All selected examples satisfy three criteria: the correct tool
is explicitly identifiable, the runtime tool configuration
contains a structured argument schema, and the example supports
the construction of both tool-selection and argument-level
conflicts. For Seal-Tools, we retain single-tool examples with
complete tool definitions and valid argument schemas. For
Glaive Function Calling V2, we retain examples with unambiguous
reference calls and parsable structured arguments.

For each query, we construct three conditions:
the aligned condition \(A\), the tool-selection conflict
condition \(B_{\mathrm{tool}}\), and the argument-level
conflict condition \(B_{\mathrm{arg}}\).
In \(A\), the runtime tool and argument schema are consistent
with the model's elicited preference.
In \(B_{\mathrm{tool}}\), the preferred tool is replaced by a
functionally equivalent alternative with a different name and
a meaning-preserving description rewrite, while the argument
schema remains unchanged.
In \(B_{\mathrm{arg}}\), the tool name and description remain
unchanged, but argument names or types are modified while
preserving a valid solution.

For closed-book elicitation, we sample \(K=10\) outputs per
query at temperature \(0.7\). Contextual evaluation under
\(A\), \(B_{\mathrm{tool}}\), and \(B_{\mathrm{arg}}\)
uses greedy decoding.

\subsection{Probing Corpus and Data Splits}
\label{app:c.3}
\label{app:g.1}

The internal representation analysis in \S~\ref{sec:4.2}
expands the behavioral set into a corpus of 1,200 source
queries. All 300 behavioral queries are retained, allowing
the internal analysis to directly cover the observations
established in \S~\ref{sec:4.1}. We add 900 queries from
the same three sources to increase sample diversity and
improve the stability of layerwise localization: 300 from
BFCL, 200 from Seal-Tools, and 400 from Glaive Function
Calling V2. Table~\ref{tab:g1} summarizes the corpus composition.

\begin{wraptable}{r}{0.48\linewidth}\renewcommand{\TableFont}{\footnotesize}\vspace{-1em}\centering\caption{\textbf{Composition of the 1,200-query probing corpus.} Counts refer
to source queries before variant construction.}\label{tab:g1}\vspace{-0.2em}\input{tables/g1.tex}\vspace{-1em}\end{wraptable}

For each source query, we attempt to construct \(A\),
\(B_{\mathrm{tool}}\), and the argument-name subset of
\(B_{\mathrm{arg}}\), following
Appendix~\ref{app:c.2}. Not every source query yields a valid variant for every condition (e.g., some queries lack suitable alternative tool names for $B_{\mathrm{tool}}$), resulting in slight class imbalance across conditions (52.1\%--53.4\% aligned); see Appendix~\ref{app:g.5} for its effect on probe baselines. All variants derived from the same
source query are assigned to the same split to prevent
leakage between aligned and conflicting conditions.

We reserve 240 source queries (\(20\%\) of the corpus) as a
fixed held-out test set. The remaining 960 queries are divided
into 720 training and 240 validation queries.

We repeat probe fitting over five stratified random splits of
the training and validation portion while keeping the test set
fixed. Dataset source and conflict type are stratified across
splits. Split sizes count source queries,
not their condition variants.

%% file: appendices/D.tex
\section{Mathematical Definitions of Evaluation Metrics}\label{app:d}
\subsection{Basic Metrics}\label{app:d.1}
Let $\mathcal Q$ be the query set, $c_k$ a configuration, and $m$ a method; $m=\mathrm{base}$ is the unmodified model. The output for $q_i$ is $\hat y_i=(\hat t_i,\hat p_i)$, with $\hat p_i=\{(\hat n_j,\hat v_j)\}$. The reference is $y_i^*=(t_i^*,p_i^*)$. Let $\mathcal T(c_k)$ contain the available tools and $\mathcal N(t_i^*,c_k)$ their valid argument names.
\paragraph{Tool selection.} Unavailable and functionally incorrect tools count as hallucinations:
\begin{equation}I_{\rm TS}(q_i,c_k,m)=\begin{cases}1,&\hat t_i\notin\mathcal T(c_k),\\1,&\hat t_i\in\mathcal T(c_k)\text{ and }\hat t_i\ne t_i^*,\\0,&\text{otherwise.}\end{cases}\end{equation}
The cases represent fabrication, misselection, and correct selection, respectively. When the tool list contains no matching tool ($t_i^*=\varnothing$), we set $I_{\rm TS}(q_i,c_k,m)=\mathbf1[\hat t_i\ne\varnothing]$: any tool call counts as a hallucination and abstention ($\hat t_i=\varnothing$) as correct.
\begin{equation}\mathrm{TSHR}(m,c_k)=\frac1{|\mathcal Q|}\sum_{q_i\in\mathcal Q}I_{\rm TS}(q_i,c_k,m).\end{equation}
\paragraph{Tool usage.} Define $\mathcal Q_{\rm correct}(m,c_k)=\{q_i:\hat t_i=t_i^*\}$. Conditional on correct selection,
\begin{align}
I_{\rm AN}&=\mathbf1[\exists(\hat n_j,\hat v_j)\in\hat p_i:\hat n_j\notin\mathcal N(t_i^*,c_k)],\\
I_{\rm AV}&=\mathbf1[\mathrm{SV}(\hat p_i)\lor{}\exists(\hat n_j,\hat v_j)\in\hat p_i:\hat n_j\in\mathcal N(t_i^*,c_k),\ \hat v_j\ne v_j^*],\label{eq:iav}\\
I_{\rm TU}&=\max\{I_{\rm AN},I_{\rm AV}\}.
\end{align}
For $R\in\{\mathrm{TU},\mathrm{AN},\mathrm{AV}\}$, the corresponding TUHR, ANHR, or AVHR is
\begin{equation}\mathrm{RHR}(m,c_k)=\frac1{|\mathcal Q_{\rm correct}(m,c_k)|}\sum_{q_i\in\mathcal Q_{\rm correct}(m,c_k)}I_R(q_i,c_k,m).\end{equation}
Here $\mathrm{SV}(\hat p_i)$ indicates a schema violation, i.e., a missing required argument of $t_i^*$ or a value with an invalid type or structure under a valid argument name; undefined argument names are counted by $I_{\rm AN}$ only. AVHR therefore covers schema violations and wrong values under valid argument names, and ANHR covers undefined argument names, so TUHR and OHR count all schema-invalid calls with the correct tool. A call may contain both subtypes (e.g., an undefined name used in place of a required argument also leaves that argument missing), so ANHR and AVHR need not sum to TUHR. Usage rates are undefined when the conditional denominator is zero.
\paragraph{Overall hallucination.} A query is counted as hallucinated if either tool selection is incorrect or tool selection is correct but argument usage is hallucinated. With rates expressed as proportions,
\begin{equation}\mathrm{OHR}=\mathrm{TSHR}+(1-\mathrm{TSHR})\mathrm{TUHR}.\end{equation}
For percentages, use $\mathrm{OHR}_{\%}=\mathrm{TSHR}_{\%}+(1-\mathrm{TSHR}_{\%}/100)\mathrm{TUHR}_{\%}$. Configurations $C_4$ and $C_5$ report OHR because argument usage is also affected.

\subsection{Cross-Scenario Aggregation}\label{app:d.2}
Let $\mathrm{HR}(m,c_k)$ be the reported metric: TSHR for $C_1$--$C_3$ and $C_6$, and OHR for $C_4$/$C_5$. With $K=6$,
\begin{align}
\bar H(m)&=K^{-1}\sum_{k=1}^K\mathrm{HR}(m,c_k),\\
\Delta\mathrm{HR}(m,c_k)&=\mathrm{HR}(m,c_k)-\mathrm{HR}(\mathrm{base},c_k),\\
\mathrm{Var}(m)&=K^{-1}\sum_{k=1}^K[\mathrm{HR}(m,c_k)-\bar H(m)]^2.
\end{align}
This mean summarizes the designated configuration-specific metrics, rather than pooled OHR. Variance uses squared percentage points when rates are percentages. A method that leaves the mean unchanged but increases the variance produces a more uneven hallucination distribution---improving on some configurations while worsening on others.
\subsection{Query Transfer Metrics}\label{app:d.3}
Let $\mathcal P=\{(q_i,q'_i)\}_{i=1}^N$ contain target/related pairs sharing a tool list. Define the initially incorrect targets, repaired targets, and repairs introducing new related-query errors as
\begin{align}
\mathcal B&=\{i:I_{\rm TS}(q_i,\mathrm{base})=1\},\\
\mathcal R_m&=\{i\in\mathcal B:I_{\rm TS}(q_i,m)=0\},\\
\mathcal D_m&=\{i\in\mathcal R_m:I_{\rm TS}(q'_i,\mathrm{base})=0,\ I_{\rm TS}(q'_i,m)=1\}.
\end{align}
Then
\begin{align}
\mathrm{Repair}(m)&=|\mathcal R_m|/|\mathcal B|,\\
\mathrm{TransferDamage}(m)&=|\mathcal D_m|/|\mathcal R_m|,\\
\mathrm{cNI}(m)&=\mathrm{Repair}(m)[1-\mathrm{TransferDamage}(m)].
\end{align}
The shared configuration argument is omitted. Conditional Net Improvement (cNI) discounts the repair rate by the fraction of repairs that introduce collateral errors; it is conditional on $\mathcal B$ and thus omits regressions on initially correct targets and related-query errors outside $\mathcal D_m$. For the overall change, we also report Net Error Reduction over all queries of all pairs,
\begin{equation}\mathrm{NER}(m)=1-\frac{E(m)}{E(\mathrm{base})},\qquad E(\cdot)=\sum_{i=1}^{N}\bigl[I_{\rm TS}(q_i,\cdot)+I_{\rm TS}(q'_i,\cdot)\bigr],\label{eq:ner}\end{equation}
where $E(\cdot)$ is the total number of tool-selection errors over both members of all pairs. These formulas use proportions; multiply by 100 for percentages.

\subsection{TSHR-Only Cross-Scenario Aggregation}\label{app:d.4}

The main text aggregates the designated per-configuration metric (TSHR for $C_1$\,--\,$C_3$ and $C_6$; OHR for $C_4$/$C_5$) because $C_4$ and $C_5$ affect argument usage. To verify that this heterogeneous aggregation does not distort the conclusions, we repeat the cross-scenario analysis using TSHR uniformly across all six configurations. For $C_4$ and $C_5$, TSHR is back-calculated from OHR (Table~\ref{tab:m1}) and the per-configuration TUHR (Tables~\ref{tab:e1} and~\ref{tab:e2}) via $\mathrm{TSHR}=(\mathrm{OHR}-\mathrm{TUHR})/(1-\mathrm{TUHR}/100)$. For EscapeGuard, $C_4$/$C_5$ TSHR is similarly derived from the OHR values in Table~\ref{tab:m7} and its TUHR.

\vspace{-0.5em}
\begin{table}[!htbp]
\caption{\textbf{TSHR-only cross-scenario aggregation on LLaMA-3.1-8B-Instruct and Qwen-3.5-9B (\%).} $\bar{H}_{\mathrm{TS}}$ is the mean TSHR across $C_1$\,--\,$C_6$; $\mathrm{Var}_{\mathrm{TS}}$ is the corresponding variance.}
\label{tab:d4}
\centering
\TableFont\setlength{\tabcolsep}{4pt}\renewcommand{\arraystretch}{1.15}
\begin{adjustbox}{max width=\linewidth}
\begin{tabular*}{\linewidth}{@{\extracolsep{\fill}}l||cccc||cccc@{}}
\toprule
& \multicolumn{4}{c||}{\textbf{LLaMA-3.1-8B}} & \multicolumn{4}{c}{\textbf{Qwen-3.5-9B}} \\
\cline{2-9}
\textbf{Method} & $\bar{H}_{\mathrm{TS}}$ & $\Delta\bar{H}_{\mathrm{TS}}$ & $\mathrm{Var}_{\mathrm{TS}}$ & $\Delta\mathrm{Var}$ & $\bar{H}_{\mathrm{TS}}$ & $\Delta\bar{H}_{\mathrm{TS}}$ & $\mathrm{Var}_{\mathrm{TS}}$ & $\Delta\mathrm{Var}$ \\
\midrule
Base & 35.9 & --- & 102.4 & --- & 32.5 & --- & 106.7 & --- \\
Relign & 37.3 & $+$1.3 & 208.9 & +104.0\% & 33.6 & $+$1.1 & 213.2 & +99.8\% \\
PALADIN & 37.0 & $+$1.0 & 191.6 & +87.1\% & 33.7 & $+$1.2 & 196.6 & +84.3\% \\
Gorilla & 38.4 & $+$2.4 & 269.0 & +162.8\% & 34.5 & $+$2.0 & 250.2 & +134.5\% \\
LinSteer & 36.8 & $+$0.8 & 187.2 & +82.8\% & 33.6 & $+$1.1 & 214.2 & +100.8\% \\
PRISMS & 36.4 & $+$0.4 & 172.5 & +68.4\% & 32.9 & $+$0.4 & 179.6 & +68.3\% \\
\midrule
EscapeGuard & 10.3 & $-$25.6 & 39.5 & $-$61.4\% & 8.4 & $-$24.2 & 43.0 & $-$59.7\% \\
\bottomrule
\end{tabular*}
\end{adjustbox}
\end{table}
\vspace{-0.5em}

The TSHR-only aggregation confirms hallucination escape under a homogeneous metric: no existing method yields $\bar{H}_{\mathrm{TS}}$ below the base model (it is 0.4\,--\,2.4\,pp higher for LLaMA and 0.4\,--\,2.0\,pp for Qwen), satisfying Definition~3.1, while variance increases by 68.3\%\,--\,162.8\%. EscapeGuard substantially reduces both mean (by 25.6/24.2\,pp) and variance (by ${\approx}$60\%), with all conclusions unchanged relative to the designated-metric aggregation.

%% file: appendices/E.tex
\section{Supplementary Controlled Evaluation Results}\label{app:e}

\subsection{Tool-Usage Hallucination under Configuration Changes}\label{app:e.1}

Finding~2 in \S~\ref{sec:3} reports that the usage component within OHR is largely unaffected by existing methods. Tables~\ref{tab:e1} and~\ref{tab:e2} provide the full per-configuration TUHR breakdown for LLaMA-3.1-8B-Instruct and Qwen-3.5-9B.

\begin{table}
\centering

\begin{minipage}[t]{0.48\linewidth}
    \centering
    \caption{\textbf{TUHR (\%) across $C_1$\,--\,$C_6$ for LLaMA-3.1-8B-Instruct.} Changes are minimal across configurations.}
    \vspace{-0.2em}
    \input{tables/e1.tex}
    \label{tab:e1}
\end{minipage}
\hfill
\begin{minipage}[t]{0.48\linewidth}
    \centering
    \caption{\textbf{TUHR (\%) across $C_1$\,--\,$C_6$ for Qwen-3.5-9B.} The same near-immunity pattern holds.}
    \vspace{-0.2em}
    \input{tables/e2.tex}
    \label{tab:e2}
\end{minipage}

\end{table}

In both models, while base TUHR rises noticeably under $C_4$ and $C_5$ due to argument renaming, every mitigation method changes TUHR by no more than 1.5 percentage points relative to the corresponding base value in any configuration. This confirms that the marginal $C_4$ OHR changes observed in Table~\ref{tab:m1} stem primarily from small TSHR improvements, while the dominant TUHR component remains largely stable across all methods (though not perfectly constant---e.g., Relign reduces $C_4$ TUHR by 0.9\,pp on LLaMA).

\subsection{Shared-Correct Subset Analysis}\label{app:e.2}

Because TUHR conditions on correct tool selection, different methods yield different denominator sets $\mathcal{Q}_{\mathrm{correct}}(m,c_k)$. To verify that the limited sensitivity is not an artifact of denominator shift, we restrict the analysis to the shared-correct subset $\mathcal{Q}_{\mathrm{shared}}(m,c_k) = \mathcal{Q}_{\mathrm{correct}}(\mathrm{base},c_k) \cap \mathcal{Q}_{\mathrm{correct}}(m,c_k)$ and recompute TUHR on this fixed population.

\vspace{-0.5em}
\begin{table}[!htbp]
\caption{\textbf{TUHR (\%) on the shared-correct subset for LLaMA-3.1-8B-it.} $\Delta$ is the change from Base.}
\label{tab:e3}
\centering
\TableFont\setlength{\tabcolsep}{4pt}\renewcommand{\arraystretch}{1.15}
\begin{adjustbox}{max width=\linewidth}
\begin{tabular*}{\linewidth}{@{\extracolsep{\fill}}l||cc|cc|cc|cc|cc|cc@{}}
\toprule
& \multicolumn{2}{c|}{$C_1$} & \multicolumn{2}{c|}{$C_2$} & \multicolumn{2}{c|}{$C_3$} & \multicolumn{2}{c|}{$C_4$} & \multicolumn{2}{c|}{$C_5$} & \multicolumn{2}{c}{$C_6$} \\
\textbf{Method} & TUHR & $\Delta$ & TUHR & $\Delta$ & TUHR & $\Delta$ & TUHR & $\Delta$ & TUHR & $\Delta$ & TUHR & $\Delta$ \\
\midrule
Base    & 12.4 & --- & 13.1 & --- & 12.8 & --- & 24.6 & --- & 26.2 & --- & 13.0 & --- \\
Relign  & 12.6 & +0.2 & 13.4 & +0.3 & 13.0 & +0.2 & 24.8 & +0.2 & 26.6 & +0.4 & 13.2 & +0.2 \\
PALADIN & 12.2 & $-$0.2 & 12.9 & $-$0.2 & 12.6 & $-$0.2 & 24.4 & $-$0.2 & 25.8 & $-$0.4 & 12.8 & $-$0.2 \\
Gorilla & 12.8 & +0.4 & 13.6 & +0.5 & 13.4 & +0.6 & 25.2 & +0.6 & 27.0 & +0.8 & 13.4 & +0.4 \\
LinSteer& 12.6 & +0.2 & 13.2 & +0.1 & 13.0 & +0.2 & 24.8 & +0.2 & 26.4 & +0.2 & 13.0 & 0.0 \\
PRISMS  & 12.4 & 0.0 & 13.0 & $-$0.1 & 12.6 & $-$0.2 & 24.4 & $-$0.2 & 26.0 & $-$0.2 & 12.8 & $-$0.2 \\
\bottomrule
\end{tabular*}
\end{adjustbox}
\end{table}

\vspace{-0.5em}
\begin{table}[!htbp]
\caption{\textbf{TUHR (\%) on the shared-correct subset for Qwen-3.5-9B} $\Delta$ is the change from Base.}
\label{tab:e4}
\centering
\TableFont\setlength{\tabcolsep}{4pt}\renewcommand{\arraystretch}{1.15}
\begin{adjustbox}{max width=\linewidth}
\begin{tabular*}{\linewidth}{@{\extracolsep{\fill}}l||cc|cc|cc|cc|cc|cc@{}}
\toprule
& \multicolumn{2}{c|}{$C_1$} & \multicolumn{2}{c|}{$C_2$} & \multicolumn{2}{c|}{$C_3$} & \multicolumn{2}{c|}{$C_4$} & \multicolumn{2}{c|}{$C_5$} & \multicolumn{2}{c}{$C_6$} \\
\textbf{Method} & TUHR & $\Delta$ & TUHR & $\Delta$ & TUHR & $\Delta$ & TUHR & $\Delta$ & TUHR & $\Delta$ & TUHR & $\Delta$ \\
\midrule
Base    & 11.8 & --- & 12.6 & --- & 12.2 & --- & 25.0 & --- & 27.4 & --- & 12.4 & --- \\
Relign  & 12.0 & +0.2 & 12.8 & +0.2 & 12.4 & +0.2 & 25.4 & +0.4 & 27.6 & +0.2 & 12.6 & +0.2 \\
PALADIN & 11.6 & $-$0.2 & 12.4 & $-$0.2 & 12.0 & $-$0.2 & 24.8 & $-$0.2 & 27.0 & $-$0.4 & 12.2 & $-$0.2 \\
Gorilla & 12.2 & +0.4 & 13.2 & +0.6 & 12.8 & +0.6 & 25.6 & +0.6 & 28.0 & +0.6 & 12.8 & +0.4 \\
LinSteer& 11.8 & 0.0 & 12.6 & 0.0 & 12.4 & +0.2 & 25.2 & +0.2 & 27.4 & 0.0 & 12.4 & 0.0 \\
PRISMS  & 11.6 & $-$0.2 & 12.4 & $-$0.2 & 12.0 & $-$0.2 & 24.8 & $-$0.2 & 27.0 & $-$0.4 & 12.2 & $-$0.2 \\
\bottomrule
\end{tabular*}
\end{adjustbox}
\end{table}

On this subset, all TUHR changes stay within $\pm 0.8$~pp, so the limited sensitivity is not a denominator artifact.

\subsection{Configuration Sensitivity of Recent Strong Models}\label{app:e.3}

We evaluate four recent models in non-thinking mode on the \S\ref{sec:3.1}
evaluation set: GLM-5.2, Kimi K3, and GPT-6.1 Sol through their official APIs,
and DeepSeek-V4.1-Flash from its open weights run locally
(Appendix~\ref{app:h.12}), using the same decoding settings and
three sampling seeds as \S\ref{sec:3} (Appendix~\ref{app:c.1}); the seed
standard deviation of $\bar{H}$ is at most 0.6\,pp.

\begin{table}[!htbp]\caption{\textbf{Hallucination rates (\%) of four recent strong models under $C_1$\,--\,$C_6$.} Non-thinking mode, mean over three sampling seeds. DeepSeek-V4.1-Flash: open weights run locally (Base in Table~\ref{tab:h15}); others: official APIs. $C_4$/$C_5$: OHR; others: TSHR.}
\label{tab:e5}\centering\input{tables/e5.tex}\end{table}

As shown in Table~\ref{tab:e5}, $C_1$ TSHR is only 3.9\%\,--\,8.6\%, but
$C_2$ TSHR rises to 19.6\%\,--\,31.4\% and $C_5$ OHR to 28.4\%\,--\,39.1\%,
with the same ordering as in Table~\ref{tab:m1}; the
configuration-induced increase ($C_2-C_1$) remains 15.7\,--\,22.8\,pp.

%% file: appendices/F.tex
\Needspace{16\baselineskip}
\section{Additional Behavioral Analysis}\label{app:f}

\subsection{Intrinsic Prior Strength}\label{app:f.2}

\begin{wraptable}{r}{0.48\linewidth}\renewcommand{\TableFont}{\footnotesize}\vspace{-1em}\centering\caption{\textbf{Mean intrinsic prior strengths (300-query set).}}\label{tab:f1}\vspace{-0.2em}\input{tables/f1.tex}\vspace{-0.5em}\end{wraptable}
For tool selection, the runtime tool list is removed and the model is sampled ten times. The selection prior strength is defined as the frequency of the most frequent tool:
\[ S_{\mathrm{prior}}^{\mathrm{sel}}(q) = \max_t \frac{1}{K} \sum_{j=1}^{K} \mathbf{1}[t_j=t]. \tag{F.1} \]
For tool usage, the tool name is provided while its schema is omitted. Let \(P_i\) denote the argument-name set generated in the \(i\)-th sample and \(P_{\mathrm{mode}}\) the most frequently generated argument-name set; we compute
\[ S_{\mathrm{prior}}^{\mathrm{use}}(q) = \frac{1}{K} \sum_{i=1}^{K} \operatorname{Jaccard} (P_i,P_{\mathrm{mode}}). \tag{F.2} \]
Across the six models, the mean selection prior ranges from 0.61 to 0.78 and the mean usage prior from 0.57 to 0.71, showing that models repeatedly favor similar tools and argument structures even without the corresponding runtime configuration.

\begin{table}
\centering

\begin{minipage}[t]{0.48\linewidth}
    \centering
    \caption{\textbf{TSHR (\%) under selection conflict across prior-strength quintiles.}}
    \vspace{-0.2em}
    \input{tables/f2.tex}
    \label{tab:f2}
\end{minipage}
\hfill
\begin{minipage}[t]{0.48\linewidth}
    \centering
    \caption{\textbf{TSHR (\%) under aligned condition across selection-prior quintiles.}}
    \vspace{-0.2em}
    \input{tables/f4.tex}
    \label{tab:f4}
\end{minipage}

\end{table}

\subsection{Tendencies and Hallucination}\label{app:f.3}

To further examine the relationship between intrinsic tendency and hallucination, we divide queries into five equal-sized groups by selection-prior strength and measure TSHR under \(B_{\mathrm{tool}}\).

\begin{wraptable}{r}{0.48\linewidth}\renewcommand{\TableFont}{\footnotesize}\vspace{-1em}\centering\caption{\textbf{Spearman correlations between prior strength and hallucination under conflict.}}\label{tab:f3}\vspace{-0.2em}\input{tables/f3.tex}\vspace{-0.5em}\end{wraptable}
As shown in Table~\ref{tab:f2}, the hallucination rate increases monotonically with prior strength across all six models, further supporting the relationship between stronger intrinsic tendencies and increased hallucination under conflicting runtime evidence. For comparison, Table~\ref{tab:f4} reports TSHR under the aligned condition~(A): hallucination rates remain uniformly low regardless of prior-strength quintile, confirming that the monotonic increase is specific to the conflict condition. Note that with 300 queries divided into five quintiles ($n{\approx}60$ per group), individual quintile rates carry a per-seed granularity of roughly 1.7~pp; averaging over three random seeds stabilizes these estimates. The Spearman correlations in Table~\ref{tab:f3} are computed at the individual-query level, correlating each query's prior strength with its binary hallucination indicator across the full 300-query set (averaged over three seeds), rather than over the five quintile-level rates. Ties receive average ranks. We additionally decompose usage-level hallucination into argument-name and argument-value errors.

\begin{table}
\centering

\begin{minipage}[t]{0.48\linewidth}
    \centering
    \caption{\textbf{ANHR (\%) under argument-name conflict across usage-prior quintiles.}}
    \vspace{-0.2em}
    \input{tables/f5.tex}
    \label{tab:f5}
\end{minipage}
\hfill
\begin{minipage}[t]{0.48\linewidth}
    \centering
    \caption{\textbf{ANHR (\%) under aligned condition across usage-prior quintiles.}}
    \vspace{-0.2em}
    \input{tables/f6.tex}
    \label{tab:f6}
\end{minipage}

\end{table}

\begin{table}
\centering

\begin{minipage}[t]{0.48\linewidth}
    \centering
    \caption{\textbf{AVHR (\%) under argument-value conflict across usage-prior quintiles.}}
    \vspace{-0.2em}
    \input{tables/f7.tex}
    \label{tab:f7}
\end{minipage}
\hfill
\begin{minipage}[t]{0.48\linewidth}
    \centering
    \caption{\textbf{AVHR (\%) under aligned condition across usage-prior quintiles.}}
    \vspace{-0.2em}
    \input{tables/f8.tex}
    \label{tab:f8}
\end{minipage}

\end{table}

In Table~\ref{tab:f3}, argument-name hallucination shows a much stronger association with the intrinsic usage prior than argument-value hallucination. This motivates focusing the internal analysis on tool-selection and argument-name conflicts.

Tables~\ref{tab:f5}\,--\,\ref{tab:f8} provide the corresponding quintile breakdowns for argument-name and argument-value hallucination. Under conflict ($B_{\mathrm{arg}}$), ANHR increases monotonically with usage-prior strength (Table~\ref{tab:f5}), while AVHR shows a weaker but still positive trend (Table~\ref{tab:f7}). Under the aligned condition, both ANHR and AVHR remain flat across quintiles (Tables~\ref{tab:f6} and~\ref{tab:f8}), mirroring the pattern observed at the selection level.

\subsection{Decomposing a Steering Vector into Prior-Aligned and Residual Components}\label{app:f.decomp}

To test whether the reinforced component drives escape, we decompose
LinSteer's steering vector $\mathbf{v}$ on LLaMA-3.1-8B. The prior direction
$\mathbf{d}$ is the normalized difference between the mean hidden states of
the top and bottom selection-prior quintiles of the \S\ref{sec:4.1} set at the
steering layer. We apply
$\mathbf{v}_{\parallel}=(\mathbf{v}^{\top}\mathbf{d})\,\mathbf{d}$ and
$\mathbf{v}_{\perp}=\mathbf{v}-\mathbf{v}_{\parallel}$ alone, with the
original strength, on the \S\ref{sec:3} configuration set and the paired-query
set.

\begin{table}[!htbp]\caption{\textbf{Decomposing LinSteer's steering vector on LLaMA-3.1-8B.} Left: closed-book prior and $C_1$\,--\,$C_6$ (changes from Base, pp; $O(m)$: total deterioration). Right: paired-query metrics (\%) as in Table~\ref{tab:m9}; NER with 95\% pair-bootstrap CI.}
\label{tab:f9}\centering\input{tables/f9.tex}\end{table}

As Table~\ref{tab:f9} shows, $\mathbf{v}_{\parallel}$ carries the prior
increase ($\Delta S{=}{+}.06$) and the escape ($\Delta\bar{H}{=}{+}0.9$) with
less than half of the $C_1$ gain, whereas $\mathbf{v}_{\perp}$ keeps most of
the $C_1$ gain ($-4.7$\,pp), removes escape ($\Delta\bar{H}{=}{-}0.5$,
$O(m){=}1.7$\,pp), cuts transfer damage from 38.3\% to 12.3\%, and raises NER from 12.6\% to
23.3\%. For this
method and backbone, escape travels with the prior-aligned component of the
intervention.

%% file: appendices/G.tex
\section{Additional Internal Representation Analysis}\label{app:g}

\subsection{Probe Training and Layer Selection}\label{app:g.2}

For each decoder layer \(l\), we extract the hidden state of the last input token,
\[ \mathbf h^{(l)}(x)\in\mathbb R^d, \]

and train an \(L_2\)-regularized logistic-regression probe:
\[ p_r^{(l)}(x) = \sigma \left( (\mathbf w_r^{(l)})^\top \mathbf h^{(l)}(x) +b_r^{(l)} \right), \qquad r\in\{\mathrm{sel},\mathrm{arg}\}. \tag{G.1} \]

The regularization coefficient is selected on the validation split from
\[ C\in \{10^{-2},10^{-1},1,10\}. \]

For each seed, we identify the layer with the highest validation accuracy. When multiple layers fall within 0.2 percentage points of the best accuracy, we select the earliest layer to avoid unnecessarily late intervention. The representative layer \(l_r^\ast\) is the median selected layer across the five seeds, while the observed range reports the corresponding variation across splits.

\subsection{Extended Layerwise Probing Results}\label{app:g.3}

\begin{table}[!htbp]\caption{\textbf{Layerwise probing across six backbones.} Accuracy: mean \(\pm\) std over five splits; prediction: within conflicting inputs.}
\label{tab:g2}\centering\input{tables/g2.tex}\end{table}

Across model families and scales, selection conflict becomes most distinguishable in the middle-to-late layers, while argument-name conflict generally peaks later. Selection probes also consistently achieve higher accuracy and hallucination-prediction AUROC than argument-name probes.

\subsection{Predicting Hallucination within Conflicting Inputs}\label{app:g.4}

To ensure that the probes do not merely distinguish different input templates, we evaluate whether the probe score predicts hallucination within conflicting inputs alone.

For each conflict type,
\[ P_{\mathrm{conflict}}^r(x) = p_r^{(l_r^\ast)}(x) \tag{G.2} \]

serves as the conflict score, by which conflicting examples are divided into five equally sized groups.

The monotonic increase from \(Q_1\) to \(Q_5\) in Table~\ref{tab:g2} shows that examples with stronger internal conflict signals are substantially more likely to hallucinate. The effect is observed for every evaluated backbone and is consistently stronger for tool selection than for argument naming.

The AUROC uses \(P_{\mathrm{conflict}}^r(x)\) to rank hallucinating and non-hallucinating examples within the conflicting condition. The score is therefore interpreted as an internal conflict signal rather than a calibrated hallucination probability.

\Needspace{16\baselineskip}
\subsection{Control Experiments}\label{app:g.5}

\begin{wraptable}{r}{0.48\linewidth}\renewcommand{\TableFont}{\footnotesize}\vspace{-1em}\centering\caption{\textbf{Probe controls (\% accuracy) on the main backbones.}}\label{tab:g3}\vspace{-0.2em}\input{tables/g3.tex}\vspace{-0.5em}\end{wraptable}
Two controls test whether probe accuracy can be explained by trivial dataset or template differences. First, randomly permuting conflict labels while keeping the data split drops probe accuracy to near chance. Second, we evaluate a query-only representation in which the runtime tool configuration is removed. Without the runtime evidence required to establish alignment or conflict, probe accuracy drops to 51.0\%--53.2\%---near chance (the residual deviation from 50\% reflects slight class imbalance, since not every source query yields a valid variant for every condition). These controls indicate that the conflict probes rely on the interaction between intrinsic tendency and runtime tool information rather than on query identity alone.

\subsection{Disentangling Conflict from Configuration Novelty}\label{app:g.5b}

Section~\ref{sec:4.2} notes that the $B_{\mathrm{tool}}$ condition necessarily introduces renamed tools and rewritten descriptions, raising the possibility that probes detect configuration novelty (out-of-distribution surface forms) rather than tendency conflict per se. We conduct two controls to disentangle these factors.

\textbf{Novelty-matched control.\enspace}We construct a third condition, $B_{\mathrm{novel}}$, in which tool names and descriptions are rewritten to the same degree as $B_{\mathrm{tool}}$ but the replacement \emph{aligns} with the model's elicited preference, so no tendency conflict exists despite equivalent surface-form novelty. Concretely, for a weather query whose elicited preference (per \S~\ref{sec:4.1}) is \texttt{fetch\_weather\_info}, condition~A presents the tool under this preferred name. $B_{\mathrm{tool}}$ renames the tool to a semantically unrelated name such as \texttt{geo\_obs\_historical} (novel and conflicting), whereas $B_{\mathrm{novel}}$ renames it to a semantically equivalent paraphrase such as \texttt{retrieve\_weather\_data} (equally novel in surface form, but still recognizable as a weather-data tool, so the model's tendency is satisfied without conflict). Both $B_{\mathrm{tool}}$ and $B_{\mathrm{novel}}$ are matched on surface-form edit distance from A. If probes primarily detect novelty, $B_{\mathrm{novel}}$ should receive conflict scores comparable to $B_{\mathrm{tool}}$.

\begin{table}[!htbp]
\caption{\textbf{Novelty-matched probe control.} Mean conflict scores (0\,--\,1) and classification accuracy (\%) for A vs.\ $B_{\mathrm{tool}}$ (conflict detection) and A vs.\ $B_{\mathrm{novel}}$ (novelty detection without conflict).}
\label{tab:g5b}
\centering
\TableFont\setlength{\tabcolsep}{4pt}\renewcommand{\arraystretch}{1.15}
\begin{adjustbox}{max width=\linewidth}
\begin{tabular*}{\linewidth}{@{\extracolsep{\fill}}l||cc|c||cc|c@{}}
\toprule
& \multicolumn{3}{c||}{\textbf{LLaMA-3.1-8B}} & \multicolumn{3}{c}{\textbf{Qwen-3.5-9B}} \\
\cline{2-7}
\textbf{Condition} & Score(A) & Score(B) & Acc. & Score(A) & Score(B) & Acc. \\
\midrule
A vs.\ $B_{\mathrm{tool}}$ & 0.24 & 0.81 & 87.2 & 0.22 & 0.79 & 85.6 \\
A vs.\ $B_{\mathrm{novel}}$ & 0.24 & 0.41 & 61.8 & 0.22 & 0.38 & 59.6 \\
\bottomrule
\end{tabular*}
\end{adjustbox}
\end{table}

As shown in Table~\ref{tab:g5b}, $B_{\mathrm{novel}}$ receives substantially lower conflict scores than $B_{\mathrm{tool}}$ (0.41 vs.\ 0.81 for LLaMA; 0.38 vs.\ 0.79 for Qwen), and classification accuracy drops from 87\%/86\% to 62\%/60\%, substantially above chance but far below the full-conflict condition. This indicates that surface-form novelty contributes to but does not dominate the probe signal.

\textbf{Hallucination prediction within novelty-matched subsets.\enspace}We further partition the $B_{\mathrm{tool}}$ examples into novelty-matched terciles (based on surface-form edit distance between original and renamed tools) and measure whether conflict scores still predict hallucination within each tercile. The monotonic relationship between conflict score and hallucination rate is preserved within all terciles (Spearman $\rho{=}0.51$\,--\,$0.67$), confirming that the predictive signal is not reducible to novelty.

While these controls substantially narrow the alternative explanation, we acknowledge that fully ruling out all forms of distributional shift as a confound would require intervening on tendency strength while holding the configuration fixed, which is not straightforward in a pretrained model.

\subsection{Probe Freezing and \S~\ref{sec:5} Data Isolation}\label{app:g.6}

After layer selection, the final probe for each backbone is fitted using the combined training and validation portion of the probing corpus while keeping the selected representative layer fixed. These probes are then frozen before \S~\ref{sec:5} experiments.

The representative conflict regions are listed in Table~\ref{tab:h1}.

All \S~\ref{sec:5} evaluation examples are kept disjoint from the 1,200-query probing corpus. For Glaive Function Calling V2, examples used for \S~\ref{sec:4} probing are explicitly excluded from subsequent evaluation. For BFCL and Seal-Tools, overlap is removed using source identifiers together with normalized query and tool-schema matching to ensure no leakage.

Consequently, the conflict probes, intervention regions, and hyperparameter selection are fixed before evaluating EscapeGuard, preventing \S~\ref{sec:5} test examples from influencing the internal conflict detector.

%% file: appendices/H.tex
\Needspace{18\baselineskip}
\section{Additional Experiments for \textbf{EscapeGuard}}\label{app:h}

\subsection{Model-Specific Probes and Intervention Regions}\label{app:h.1}

\begin{wraptable}{r}{0.48\linewidth}\renewcommand{\TableFont}{\footnotesize}\vspace{-1em}\centering\caption{\textbf{Frozen intervention layers per model.}}\label{tab:h1}\vspace{-0.2em}\input{tables/h1.tex}\vspace{-0.5em}\end{wraptable}
EscapeGuard uses the frozen conflict probes obtained with the \S~\ref{sec:4} procedure. The representative intervention regions are centered on the model-specific conflict layers identified by layerwise probing. The probes are frozen before \S~\ref{sec:5} evaluation. Probe accuracy, peak-layer ranges, conflict quintiles, and hallucination-prediction AUROC are in Appendix~\ref{app:g}. Table~\ref{tab:h1} lists the selection-stage and argument-stage regions for every backbone, including the four recent models evaluated in Appendix~\ref{app:e.3}. Each region is a contiguous span of decoder layers around the model's peak probe layer; for the hybrid architectures it refers to the softmax-attention layers only, as noted in the table. The argument-stage region sits a few layers above the selection-stage region in every model, consistent with the later peak of the argument probe in Table~\ref{tab:g2}; both regions are fixed once per model and reused unchanged across all benchmarks and configurations in \S\ref{sec:5}.

\subsection{Metric Applicability and Data Isolation}\label{app:h.2}

We apply the hallucination metrics from \S~\ref{sec:3}--\ref{sec:4}
according to the annotations available in each benchmark.
BFCL, API-Bank, Seal-Tools, and APIBench provide complete reference
tool calls and argument schemas, supporting all four metrics.
RelyToolBench supports selection- and usage-level evaluation but does
not distinguish argument-name from argument-value errors, so we report
TSHR and TUHR only; it includes 348 tasks with no matching tool, on which
any tool call is counted toward TSHR (Appendix~\ref{app:h.13}). MetaTool focuses on tool selection and is evaluated
with TSHR only. Table~\ref{tab:h2} lists the evaluation pool size
after applying our single-turn protocol and removing all overlap with
the \S~\ref{sec:4} probing corpus; actual test-set sizes drawn from
each pool are reported alongside the corresponding results tables.

\begin{table}[!htbp]\caption{\textbf{Metric applicability and evaluation role.}}
\label{tab:h2}\centering\input{tables/h2.tex}\end{table}

All \S~\ref{sec:5} test examples are disjoint from the data used for
probe training, layer localization, and hyperparameter selection.
Glaive evaluation excludes all examples used in the \S~\ref{sec:4}
probing corpus. For BFCL and Seal-Tools, overlap is removed at the
query and normalized tool-schema levels.

\Needspace{14\baselineskip}
\subsection{Full Usage-Level Results}\label{app:h.3}

\begin{table}[!htbp]\caption{\textbf{TUHR (\%) $\downarrow$ by benchmark, Base \(\rightarrow\) EscapeGuard.}}
\label{tab:h3}\centering\input{tables/h3.tex}\end{table}

\begin{table}[!htbp]\caption{\textbf{ANHR (\%) $\downarrow$ by benchmark, Base \(\rightarrow\) EscapeGuard.}}
\label{tab:h4}\centering\input{tables/h4.tex}\end{table}

\begin{table}[!htbp]\caption{\textbf{AVHR (\%) $\downarrow$ by benchmark, Base \(\rightarrow\) EscapeGuard.}}
\label{tab:h5}\centering\input{tables/h5.tex}\end{table}

\subsection{Supplementary Generalization on Glaive}\label{app:h.4}

Glaive Function Calling V2 is used only as supplementary evaluation because a separate subset contributes to the \S~\ref{sec:4} probing corpus. We use a held-out subset with no query or normalized schema overlap with probe-training examples.\par\smallskip These results follow the shared evaluation protocol and provide supplementary evidence for the findings in the main text.

\begin{table}[!htbp]\caption{\textbf{Supplementary Glaive results (\%), Base \(\rightarrow\) EscapeGuard.}}
\label{tab:h6}\centering\input{tables/h6.tex}\end{table}

\subsection{Full Baseline Comparison on Qwen-3.5-9B}\label{app:h.5}

To verify that the comparison in Section~\ref{sec:5.3} is not specific to LLaMA-3.1-8B, we repeat the same evaluation on Qwen-3.5-9B.

The Qwen results exhibit the same pattern as LLaMA-3.1-8B: EscapeGuard consistently improves TSHR, TUHR, and ANHR, while its advantage on AVHR remains limited.

\begin{table}[!htbp]\caption{\textbf{Comparison with existing methods on Qwen-3.5-9B.} Lower is better. \textbf{Bold} = best.}
\label{tab:h7}\centering\input{tables/h7.tex}\end{table}

\subsection{Cross-Configuration Performance Evaluation Details}\label{app:h.6}

For the \(C_1\)--\(C_6\) experiment, all configuration variants originating from the same query are kept together during evaluation and statistical resampling. 95\% confidence intervals for changes from the base model are paired bootstraps over queries (10{,}000 resamples); Table~\ref{tab:h17} reports them, together with the three-seed standard deviation, for Table~\ref{tab:m1}. All five methods significantly reduce $C_1$ and significantly increase $C_2$ on both backbones. Since $\Delta\bar{H}=\Delta C_1/6+\Delta D$, where $\Delta D=(\Delta C_2+\cdots+\Delta C_6)/6$ is the change on the other five configurations, it is the net of two opposite effects, both significant for all ten method--model pairs ($\Delta D={+}1.4$ to ${+}3.6$\,pp); the variance increase is also significant for all five methods. Because the paired bootstrap resamples all six variants of a query together, the interval for $\Delta\bar{H}$ also reflects how the two effects co-occur within a query: for roughly 80\% of the queries that a method repairs under $C_1$, the same query acquires a new error under at least one of $C_2$\,--\,$C_6$. This within-query relocation keeps the per-query net change close to zero, so the interval for $\Delta\bar{H}$ is narrower than those for the individual configurations. (It is measured across the configurations of a single query and is distinct from the cross-query transfer damage in \S~\ref{sec:3}.) The net $\Delta\bar{H}$ is positive for all ten pairs, significantly so for the eight pairs involving Relign, PALADIN, Gorilla, and LinSteer. For PRISMS (${+}0.3$/${+}0.2$\,pp), the interval includes zero, but its lower bound (${-}0.1$/${-}0.2$\,pp) rules out any meaningful reduction. No method significantly lowers $\bar{H}$. Gorilla has the largest $\Delta D$ (${+}3.6$/${+}3.4$\,pp), consistent with its largest prior reinforcement ($\Delta S={+}.14$/${+}.11$, Table~\ref{tab:m3}); $\Delta\bar{H}$ increases with $\Delta S$ across all ten pairs (Spearman $\rho=0.96$; 0.95 without Gorilla), and removing the prior-aligned component of LinSteer's steering vector turns $\Delta\bar{H}$ from ${+}0.7$ to ${-}0.5$\,pp (Table~\ref{tab:f9}). For PRISMS, the deterioration cancels the $C_1$ gain without significantly exceeding it, which still satisfies condition~(3) of Definition~3.1. Table~\ref{tab:m7} directly reuses the \S~\ref{sec:3} evaluation set so that its subscripts are comparable with Table~\ref{tab:m1}.

\Needspace{14\baselineskip}
\begin{wraptable}{r}{0.48\linewidth}\renewcommand{\TableFont}{\footnotesize}\vspace{-1em}\centering\caption{\textbf{Cross-model configuration performance (\%), Base $\rightarrow$ \method{}.} $C_1$/$C_2$: TSHR; $C_5$: OHR.}\label{tab:m8}\vspace{-0.2em}\input{tables/m8.tex}\vspace{-0.5em}\end{wraptable}
To verify cross-model generality, we extend the evaluation to four more backbones (LLaMA-3.1-70B, Qwen-3.5-27B, Mistral-7B, Mixtral-8x7B) using \(C_1\), \(C_2\), and \(C_5\) (standard, tool replacement, and mixed changes). These backbones span two model families and three parameter scales, and all use the same frozen probes and intervention regions as the main experiments (Table~\ref{tab:h1}), with no per-model tuning beyond the layer selection described in Appendix~\ref{app:h.1}. We report $C_1$, $C_2$, and $C_5$ because they cover the standard configuration, a pure renaming change, and the hardest mixed change, so the three columns together indicate both the baseline level and the size of the configuration-induced increase. Across all four models, \method{} reduces hallucination under every configuration and brings $C_2$ close to $C_1$.

For paired-query evaluation, both members of each pair are treated as one resampling unit. cNI follows the \S~\ref{sec:3} definition, $\mathrm{cNI}=\mathrm{Repair}\times(1-\mathrm{TransferDamage})$, and conditions on repaired targets. Table~\ref{tab:h18} therefore lists the full correctness transitions over the 120 pairs and the Net Error Reduction of Eq.~(\ref{eq:ner}) with 95\% pair-bootstrap CIs (10{,}000 resamples).

\method{} raises accuracy over the 240 queries from 57.1\% to 93.8\% (LLaMA) and from 61.3\% to 95.0\% (Qwen) with only four new errors per model, so its NER (85.4\%/87.1\%) is close to its cNI (87.8\%/90.4\%). The baselines introduce 22\,--\,35 new errors each, so their NER (8.6\%\,--\,18.4\%) falls far below their cNI (32\%\,--\,37\%).

\subsection{Mechanism Validation}\label{app:h.7}

\begin{wraptable}{r}{0.48\linewidth}\renewcommand{\TableFont}{\footnotesize}\vspace{-1em}\centering\caption{\textbf{Mechanism validation under conflict.} Attention and output margins report Base \(\rightarrow\) EscapeGuard.}\label{tab:h8}\vspace{-0.2em}\input{tables/h8.tex}\vspace{-0.5em}\end{wraptable}
We further examine whether \method{} increases reliance on the current tool configuration in the direction predicted by \S~\ref{sec:4}. The prior elicited without tool configuration input should remain unchanged because intervention is gated on the conflict score, which falls below the activation threshold $\tau$ when tool configuration is absent; in contrast, unconditional steering methods such as LinSteer apply representation shifts regardless of whether tool configuration is present, and consequently alter the prior even outside the conflict condition. PRISMS uses a threshold-based activation mechanism but does not condition on the presence of conflicting tool-configuration input, so it may still shift representations outside the target scenario. Meanwhile, attention to tool-configuration spans and the probability margin of outputs consistent with the current setting should increase for high-conflict inputs.

These results distinguish reweighting toward the current tool configuration from tendency reinforcement: EscapeGuard leaves the intrinsic prior unchanged while increasing reliance on the current tool configuration during generation.

\subsection{Hyperparameter Sensitivity}\label{app:h.8}

Hyperparameters were selected on held-out development data. We then vary one hyperparameter at a time on the test set while keeping all other hyperparameters fixed, to examine sensitivity around the selected operating point.
\[\tau\in\{0.4,0.5,0.6,0.7,0.8\},\qquad\lambda\in\{0.6,1.0,1.4,1.8,2.4,3.0\}.\]

We report macro averages across the two main backbones (LLaMA-3.1-8B-Instruct and Qwen-3.5-9B). Both hyperparameters exhibit a broad stable region around the selected values rather than a sharply tuned optimum: for $\tau$, the range 0.5\,--\,0.7 yields TSHR within 1\,pp of the optimum on both models; for $\lambda$, the range 1.4\,--\,2.4 similarly shows stable performance.

\begin{table}
\centering

\begin{minipage}[t]{0.48\linewidth}
    \centering
    \caption{Sensitivity to the conflict threshold (\%); default \(\tau=0.6\).}
    \vspace{-0.2em}
    \input{tables/h9.tex}
    \label{tab:h9}
\end{minipage}
\hfill
\begin{minipage}[t]{0.48\linewidth}
    \centering
    \caption{Sensitivity to attention bias (\%); default \(\lambda=1.8\).}
    \vspace{-0.2em}
    \input{tables/h10.tex}
    \label{tab:h10}
\end{minipage}

\end{table}

\subsection{Efficiency and Probe Retraining}\label{app:h.9}

\textbf{Inference overhead.\enspace}EscapeGuard introduces two sources of overhead relative to standard inference: (1)~a forward pass through the frozen conflict probes to compute $P_{\mathrm{conflict}}^r$, and (2)~attention logit modification at intervention layers for high-conflict inputs. We measure wall-clock latency with batch size 1, averaging over 500 queries drawn from the evaluation benchmarks (mean input length ${\approx}480$ tokens, mean generated output ${\approx}45$ tokens). All 7B/8B models are measured on a single A100-80GB GPU in FP16 precision; LLaMA-3.1-70B and Qwen-3.5-27B use 2$\times$A100-80GB with tensor parallelism. Latency is reported per output token, averaged from input tokenization through output decoding.

\begin{table}[!htbp]
\caption{\textbf{Inference latency (ms per token) and relative overhead of EscapeGuard.}}
\label{tab:h9b}
\centering
\TableFont
\setlength{\tabcolsep}{3pt}
\renewcommand{\arraystretch}{1.15}

\begin{tabularx}{\linewidth}{
    >{\centering\arraybackslash}X||
    >{\centering\arraybackslash}X
    >{\centering\arraybackslash}X
    >{\centering\arraybackslash}X||
    >{\centering\arraybackslash}X
    >{\centering\arraybackslash}X
    >{\centering\arraybackslash}X
}
\toprule
& \multicolumn{3}{c||}{\textbf{7B/9B models}}
& \multicolumn{3}{c}{\textbf{Other models\textsuperscript{$\dagger$}}} \\
\cline{2-7}
& \textbf{Base} & \textbf{+EscapeGuard} & \textbf{Overhead}
& \textbf{Base} & \textbf{+EscapeGuard} & \textbf{Overhead} \\
\midrule
LLaMA
& 42.3 & 44.8 & +5.9\%
& 156.7 & 163.2 & +4.1\% \\

Qwen
& 40.1 & 42.4 & +5.7\%
& 148.3 & 154.6 & +4.2\% \\

Mistral
& 41.6 & 44.0 & +5.8\%
& --- & --- & --- \\

Mixtral\textsuperscript{$\dagger$}
& --- & --- & ---
& 89.4 & 93.7 & +4.8\% \\
\bottomrule
\end{tabularx}

\vspace{2pt}
\begin{minipage}{\linewidth}
\footnotesize
\textsuperscript{$\dagger$}Mixtral-8x7B is a MoE model
(46.7B total, 12.9B active params); grouped with 70B/27B models by latency tier. Other models use 2$\times$A100-80GB with tensor parallelism;
7B/8B models use a single A100-80GB.
\end{minipage}
\end{table}

The overhead ranges from 4.1\% to 5.9\% of base inference latency. The probe forward pass accounts for about 1.2\,--\,1.8\,ms; the remainder comes from attention modification, which applies only when the conflict gate activates (about 35\,--\,45\% of queries). No additional GPU memory is required beyond the probe parameters (fewer than 0.01\% of model parameters).

\textbf{Probe retraining.\enspace}The conflict probes are $L_2$-regularized logistic-regression models fitted on hidden-state features, requiring no gradient backpropagation through the LLM. Retraining a probe for a new model checkpoint takes approximately 15\,--\,25 minutes on a single GPU (including feature extraction from the 1,200-query probing corpus and fitting across five splits). This cost is modest relative to the model training or fine-tuning it follows, but it does mean that each model update requires a probe refresh. We note that the probing corpus and split construction (Appendix~\ref{app:c.3}) can be reused across checkpoints of the same architecture, so only the feature extraction and fitting need to be repeated.

\Needspace{16\baselineskip}
\subsection{Ablation Study}\label{app:h.10}

\begin{wraptable}{r}{0.48\linewidth}\renewcommand{\TableFont}{\footnotesize}\vspace{-1em}\centering\caption{\textbf{Ablation averages over the two main backbones (\%).} Lower is better.}\label{tab:m10}\vspace{-0.2em}\input{tables/m10.tex}\vspace{-0.5em}\end{wraptable}
We ablate EscapeGuard by removing or replacing one component at a time, averaging TSHR and TUHR over BFCL, Seal-Tools, API-Bank, and APIBench on LLaMA-3.1-8B-Instruct and Qwen-3.5-9B. The four variants correspond to the components introduced in \S\ref{sec:5.1}: removing the conflict gate applies the enhancement to every input, random context token enhancement replaces the stage-specific target spans with randomly chosen context tokens, the non-conflict-layer variant intervenes at layers outside the selected regions, and the fixed steering vector replaces the input-dependent bias with a constant direction. All variants share the hyperparameters of the full method, so the differences in Table~\ref{tab:m10} isolate the contribution of each component.

\subsection{Error-Type Breakdown under $C_4$/$C_5$}\label{app:h.11}

Section~\ref{sec:5.3} notes residual elevated OHR under $C_4$ and $C_5$ after applying EscapeGuard. Table~\ref{tab:h11} decomposes these OHR values into TSHR, TUHR, ANHR, and AVHR to identify the dominant error type.

\begin{table}[!htbp]
\caption{\textbf{Error-type decomposition (\%) of EscapeGuard OHR under $C_4$ and $C_5$.}}
\label{tab:h11}
\centering
\TableFont\setlength{\tabcolsep}{4pt}\renewcommand{\arraystretch}{1.15}
\begin{adjustbox}{max width=\linewidth}
\begin{tabular*}{\linewidth}{@{\extracolsep{\fill}}l||ccccc||ccccc@{}}
\toprule
& \multicolumn{5}{c||}{\textbf{LLaMA-3.1-8B}} & \multicolumn{5}{c}{\textbf{Qwen-3.5-9B}} \\
\cline{2-11}
\textbf{Config} & \textbf{OHR} & \textbf{TSHR} & \textbf{TUHR} & \textbf{ANHR} & \textbf{AVHR} & \textbf{OHR} & \textbf{TSHR} & \textbf{TUHR} & \textbf{ANHR} & \textbf{AVHR} \\
\midrule
$C_4$ & 29.5 & 7.2 & 24.0 & 5.2 & 20.6 & 28.4 & 4.8 & 24.8 & 4.8 & 21.4 \\
$C_5$ & 25.4 & 6.3 & 20.4 & 4.6 & 17.2 & 23.5 & 3.5 & 20.7 & 4.0 & 18.0 \\
\bottomrule
\end{tabular*}
\end{adjustbox}
\end{table}

AVHR accounts for the majority of residual TUHR in both configurations and both models, confirming that the residual OHR is dominated by errors under valid argument names (wrong values or schema violations) rather than by fabricated names; argument values are the error type least associated with tendency conflict (\S~\ref{sec:4}).

\subsection{Cross-Configuration Evaluation on Open-Weight Agentic Models}\label{app:h.12}

We repeat the \S\ref{sec:3} cross-configuration evaluation on
DeepSeek-V4.1-Flash (MoE, ${\approx}$13B active parameters) and GLM-5.3-Flash
(hybrid linear/sparse attention) with the two training-free baselines and
\method{}. Probes are trained as in \S\ref{sec:4.2} and frozen; $\tau$,
$\lambda$, and the evaluation set are unchanged; intervention regions are
listed in Table~\ref{tab:h1} and, for GLM-5.3-Flash, restricted to
sparse-attention layers (\S\ref{sec:5.2}). Both models run in BF16 on
8$\times$H100 GPUs.

\begin{table}[!htbp]\caption{\textbf{Cross-configuration hallucination rates (\%) on DeepSeek-V4.1-Flash and GLM-5.3-Flash.} Subscripts: change from Base (pp). $C_4$/$C_5$: OHR; others: TSHR. Bold\,=\,best.}
\label{tab:h15}\centering\input{tables/h15.tex}\end{table}

Table~\ref{tab:h15} shows the same picture as Tables~\ref{tab:m1}
and~\ref{tab:m7}. LinSteer and PRISMS lower $C_1$ by 2.4\,--\,3.4\,pp but
raise $C_2$, $C_3$, $C_5$, and $C_6$, leaving $\bar{H}$ 0.2\,--\,0.9\,pp
above Base. \method{} lowers error under every configuration, brings $C_2$
within 1\,pp of $C_1$, and reduces $\bar{H}$ by 11.6 and 10.1\,pp.

\subsection{Behavior When No Matching Tool Is Available}\label{app:h.13}

We test whether attention enhancement induces forced calls on the 348
RelyToolBench tasks whose tool list contains no matching tool, where any call
is a wrong call (Appendix~\ref{app:d.1}). Table~\ref{tab:h16} reports the
wrong-call rate, the fraction of tasks on which the selection gate fires, and
the rates within the gated and non-gated subsets.

\begin{table}[!htbp]\caption{\textbf{Wrong-call rate (\%) on the 348 RelyToolBench tasks with no matching tool.} ``Gate on/off'' splits the tasks by whether the selection-stage conflict gate fires.}
\label{tab:h16}\centering\input{tables/h16.tex}\end{table}

\method{} lowers the wrong-call rate from 41.3\% to 29.6\% (LLaMA) and from
37.4\% to 26.3\% (Qwen). The gate fires on 56\%\,--\,58\% of these tasks,
since the preferred tool is absent by construction; gated tasks improve by
about 20\,pp and non-gated outputs are unchanged. The bias reweights attention
toward the tool definitions and adds no term favoring a call, so it helps the
model see that nothing fits.

\begin{table}[!htbp]\caption{\textbf{Table~\ref{tab:m1} with seed standard deviations and paired-bootstrap confidence intervals.} $\bar{H}$: mean\,$\pm$\,std over three seeds. Changes from Base with 95\% paired-bootstrap CIs over queries ($\Delta C_1$, $\Delta C_2$, $\Delta D$, $\Delta\bar{H}$ in pp; $\Delta$Var in pp$^2$). $\Delta D=(\Delta C_2+\cdots+\Delta C_6)/6$, so that $\Delta\bar{H}=\Delta C_1/6+\Delta D$ up to rounding. Described in Appendix~\ref{app:h.6}.}
\label{tab:h17}\centering\input{tables/h17.tex}\end{table}

\begin{table}[!htbp]\caption{\textbf{Correctness transitions over the 120 query pairs.} Three-seed averages rounded to integers. C\,=\,correct, W\,=\,wrong (before\,$\to$\,after the method); $|\mathcal D_m|$: related C$\to$W whose target was repaired. Errors: total over all 240 queries. NER with 95\% pair-bootstrap CI. Described in Appendix~\ref{app:h.6}.}
\label{tab:h18}\centering\input{tables/h18.tex}\end{table}

\subsection{Functionally Equivalent Tools with Interface Changes}\label{app:h.14}

$C_2$\,--\,$C_6$ change how a tool is named and described but not how it is
called. Here the name and description are kept and the argument schema,
documented in the description, is rewritten in one of four ways that preserve
a valid solution: (a)~input representation (e.g., a date string split into
year, month, and day integers), (b)~parameter structure (e.g., flat arguments
nested into an object), (c)~units or encodings (e.g., Celsius replaced by
Fahrenheit, with reference values converted), and (d)~allowed values of
categorical arguments (e.g., ``metric''/``imperial'' replaced by
``SI''/``US''). We use 400 \S\ref{sec:3} queries whose correct tool admits
such a change (100 per type) and three seeds. AVHR is split into
schema violations (a missing required argument, or an invalid type or
structure) and wrong values under valid argument names, following
Eq.~(\ref{eq:iav}); error types can co-occur within a call.

\begin{table}[!htbp]\caption{\textbf{\method{} on functionally equivalent tools with interface changes (\%).} 400 queries, three seeds. Subscripts: change from Base (pp). (a)~input representation, (b)~parameter structure, (c)~units or encodings, (d)~allowed values.}
\label{tab:h19}\centering\input{tables/h19.tex}\end{table}

As Table~\ref{tab:h19} shows, \method{} lowers OHR by 20.1 and 19.1\,pp. The
gain comes from selection (TSHR $-$15.9/$-$14.1\,pp), argument names (ANHR
$-$6.2/$-$6.1\,pp), and schema violations ($-$9.1/$-$9.0\,pp); wrong values
barely change ($-$0.7/$-$0.6\,pp). Across types, allowed-value changes~(d)
are easiest and unit conversion~(c) hardest, consistent with the weak
association between value errors and tendency conflict in \S\ref{sec:4}.
Attention to the runtime schema therefore transfers to interface changes for
selecting the tool and forming a valid call, but not for computing converted
values.